\documentclass[jounral]{IEEEtran}  

\usepackage{amsmath}
\usepackage{amssymb}
\usepackage[colorinlistoftodos]{todonotes}
\usepackage{hyperref}
\usepackage{multirow}
\usepackage{booktabs}
\usepackage[colorinlistoftodos]{todonotes}
\usepackage{siunitx}
\usepackage{url}
\usepackage{import,bm}
\usepackage[super]{nth}
\usepackage{romannum}
\usepackage{accents}
\usepackage{verbatim}
\usepackage{tikz}
\usetikzlibrary{automata, positioning, arrows, shapes.geometric, arrows.meta, positioning, calc, intersections, spath3}
\usepackage{tkz-euclide}
\usepackage{subcaption}
\usepackage{graphicx}
\usepackage{cite}
\usepackage{makecell}
\usepackage{caption}
\usepackage{ragged2e}
\usepackage{soul}
\usepackage{textcomp}
\usepackage{url}
\usepackage{listings}
\usepackage{helvet}
\newcommand{\argmin}{\mathop{\mathrm{argmin}}}

\DeclareMathOperator{\sign}{sign}

\newcommand{\mt}[1]{\bm{#1}}

\title{Haptic Transparency and Interaction Force Control for a Lower-Limb Exoskeleton 
}
\author{Emek Barış Küçüktabak, Yue Wen, Sangjoon J. Kim, Matthew R. Short, Daniel~Ludvig,\\ Levi~Hargrove, Eric~J.~Perreault, Kevin M. Lynch, and Jose L. Pons

\thanks{E. B. Küçüktabak is with the Center for Robotics and Biosystems, Department of Mechanical Engineering, Northwestern University, Evanston, IL, USA and Shirley Ryan AbilityLab, Chicago, IL USA.}
\thanks{S. J. Kim and Y. Wen are with the Shirley Ryan AbilityLab and Department of Physical Medicine and Rehabilitation, Northwestern University, Chicago, IL, USA.}
\thanks{M. R. Short is with the Shirley Ryan AbilityLab and Department of Biomedical Engineering, Northwestern University, Evanston, IL, USA}
\thanks{D. Ludvig is with the Department of Biomedical Engineering, Northwestern University.}
\thanks{L. Hargrove is with the Shirley Ryan AbilityLab, Center for Robotics and Biosystems, and Department of Physical Medicine and Rehabilitation, Northwestern University.}
\thanks{E. J. Perreault is with the Department of Physical Medicine and Rehabilitation, Department of Biomedical Engineering, Shirley Ryan AbilityLab and Center for Robotics and Biosystems.}
\thanks{K. M. Lynch is with Center for Robotics and Biosystems and Department of Mechanical Engineering, Northwestern University.}
\thanks{J. Pons is with the Shirley Ryan AbilityLab, Center for Robotics and Biosystems, Department of Mechanical Engineering, Department of Biomedical Engineering, and Department of Physical Medicine and Rehabilitation, Northwestern University.}

}

\usepackage{atbegshi,picture}
\usepackage{lipsum}

\AtBeginShipoutNext{\AtBeginShipoutUpperLeft{%
  \put(\dimexpr\paperwidth-1cm\relax,-1.5cm){\makebox[0pt][r]{\framebox{Accepted for publication in the IEEE Transactions on Robotics (T-RO).}}}%
}}

\begin{document}
\maketitle
\thispagestyle{empty}
\pagenumbering{arabic}

\begin{abstract}
\textcolor{black}{
Controlling the interaction forces between a human and an exoskeleton is crucial for providing transparency or adjusting assistance or resistance levels. However, it is an open problem to control the interaction forces of lower-limb exoskeletons designed for unrestricted overground walking. For these types of exoskeletons, it is challenging to implement force/torque sensors at every contact between the user and the exoskeleton for direct force measurement. Moreover, it is important to compensate for the exoskeleton's whole-body gravitational and dynamical forces, especially for heavy lower-limb exoskeletons. Previous works either simplified the dynamic model by treating the legs as independent double pendulums, or they did not close the loop with interaction force feedback.}

\textcolor{black}{The proposed whole-exoskeleton closed-loop compensation (WECC) method calculates the interaction torques during the complete gait cycle by using whole-body dynamics and joint torque measurements on a hip-knee exoskeleton. Furthermore, it uses a constrained optimization scheme to track desired interaction torques in a closed loop while considering physical and safety constraints. We evaluated the haptic transparency and dynamic interaction torque tracking of WECC control on three subjects. We also compared the performance of WECC with a controller based on a simplified dynamic model and a passive version of the exoskeleton. 
The WECC controller results in a consistently low absolute interaction torque error  
during the whole gait cycle for both zero and nonzero desired interaction torques. In contrast, the simplified controller yields poor performance in tracking desired interaction torques during the stance phase. 
}


\end{abstract}

\section{Introduction}
\label{sec:Introduction}

Lower-limb exoskeletons are used to assist impaired individuals in their daily life and as a tool for physical rehabilitation in clinical settings~\cite{Moreno2018, 6930719}. They are also increasingly used to augment the abilities of healthy individuals~\cite{Sawicki2020}. 
In each of these applications, the exoskeleton must be capable of regulating the interaction forces and torques between the user and the exoskeleton, both for the safety of the user and the effectiveness of the control mode. 

As an example, the \emph{transparent} control mode attempts to zero the interaction forces between the user and the exoskeleton. Transparency is useful for comparison when evaluating new rehabilitation control strategies~\cite{Marchal-Crespo2019-zo} and for allowing users to move freely when wearing assistive devices~\cite{Lee2017}. Other control modes, used in rehabilitation, include ``assist as needed'' (AAN) and resistive control for robot-aided gait therapy~\cite{Hobbs2020}. With AAN, the robot provides only as much assistance force as needed to complete a given task~\cite{Baud2021}. In resistive and error-augmentation methods, force is applied to make the task more challenging, which accelerates relearning of motor tasks~\cite{Marchal-Crespo2009, Marchal-Crespo2019-zo}. 



An important part of controlling interaction forces is compensating the weight of the lower-limb exoskeleton. How the weight of the exoskeleton is transferred to the ground or the user depends on its design. There are three main categories of design:
\paragraph{Grounded trunk}
The trunk of the exoskeleton is fixed relative to a treadmill structure~\cite{colombo2000treadmill, Banala2007, Veneman2007}. Due to the grounded trunk, only the masses of the individual leg links are felt by the user, and these masses can be compensated with active control.
\paragraph{Floating-base (trunk) without feet}
These exoskeletons are designed for unrestricted overground walking and do not contact the ground~\cite{MurraySA2015Anstroke, Tsukahara2018, Mayag2022}. Partial gravity compensation can be achieved for the swing leg, but it is not possible for the exoskeleton to completely carry its own weight since there is no contact with the ground. Such exoskeletons are designed to be as light as possible, which limits their maximum assistance/resistance and battery life.
\paragraph{Floating-base with feet}
These exoskeletons are designed for unrestricted overground walking and have feet that contact the ground~\cite{Camardella2021, VantiltJ2019Model-basedmovements, Andrade2019, Andrade2021, Tu2020, Sharifi2021, Kazerooni2006}. The weight of the exoskeleton can be transferred to the ground through the exoskeleton's foot contact. While this allows for stronger actuators and larger batteries, it comes at the cost of more complicated controllers \textcolor{black}{for accurate interaction force control}.

In this paper, we focus on floating-base lower-limb exoskeletons with feet, for which controlling interaction forces remains an open problem~\cite{TRO_Special, Andrade2019}.
\textcolor{black}{Accurate interaction force control will enable natural implementations of AAN, resistive, and error-augmentation methods in overground walking therapy using rehabilitation exoskeletons. Floating-base exoskeletons with feet also allow the possibility of larger assistive forces and longer battery life.}

Due to the high torque requirements of floating-base exoskeletons with feet, high reduction ratios are used, which results in high friction and apparent inertia of motor rotors~\cite{Andrade2021Role}. Also, these exoskeletons are heavy due to powerful actuators and large batteries. Therefore, it is essential to compensate for these factors to accurately control the interaction forces and torques between the user and the exoskeleton.

Another challenge is measuring the interaction torques between the human and the exoskeleton. Grounded exoskeletons typically have a small number of contacts with the user, and interaction forces can be directly measured by force sensors implemented at the contacts~\cite{Banala2007, ZanottoD2013Improvingcuffs, vanDijkW2013Improvingwalking}. In contrast, floating-base exoskeletons have broadly distributed contacts with the user, including at the trunk and foot. \textcolor{black}{This makes it challenging to implement force/torque sensors at all contacts.}
Therefore, it is required to indirectly estimate the interaction forces and torques through other sensory information.

In this paper, we present a novel method to estimate and control user-exoskeleton interaction torques during the complete gait cycle of a floating-base hip-knee exoskeleton with feet. The controller, which is based on constrained optimization, outperforms the state of the art in achieving transparency (zero impedance) and nonzero impedances in the user-exoskeleton interaction behavior.

The context of the contribution of this paper is established by the related work below.




\subsection{Related Work}
\label{sec:Related_Work}

\begin{table*}
\centering
\caption{Selected features of the interaction force/torque control studies. Note that this paper focuses on floating-base lower-limb exoskeletons with feet. Therefore, this table does not include studies with grounded exoskeletons and floating-base lower-limb exoskeletons without feet.}
\begin{tabular}{c c c c c c c} 
\cline{1-6}
\rule{0pt}{2ex}  
  Reference & Model & \makecell{Interaction force\\ feedback} & \makecell{User\\dependent} & \makecell{Instrumentation\\ on user} & \makecell{Physical/safety \\limit implementation}  \\ \cline{1-6}
  
  \makecell{Camardella et al.,\\ 2021 \cite{Camardella2021}} & Whole-body & No & Yes & No & No \\
    \cline{1-6}
  \makecell{Vantilt et al.,\\ 2019 \cite{VantiltJ2019Model-basedmovements}} & Whole-body & No & No & No & No \\
    \cline{1-6}
  \makecell{Andrade et al.,\\ 2019-21 \cite{Andrade2019, Andrade2021}} & \makecell{Independent\\ double pendulums} &  \makecell{PD\\ controller} & No & No & No \\
    \cline{1-6}
  \makecell{Tu et al.,\\ 2020 \cite{Tu2020}} & \makecell{Independent\\ double pendulums} & \makecell{Admittance\\ controller} & No & No & No \\
    \cline{1-6}  
  \makecell{Sharifi et al.,\\ 2021 \cite{Sharifi2021}} & \makecell{Whole-body\\(learnt by NN)} & No & Yes & No & No \\
  \cline{1-6}
  \makecell{Kazerooni et al., \\ 2006 \cite{Kazerooni2006}} & \makecell{3-link for swing\\no model for stance\\(position control)} & \makecell{Sensitivity\\Amplification} & \makecell{No (transparency)\\Yes (non-zero reference)} & Yes & No \\
  \cline{1-6}
  \makecell{Our proposed\\method (WECC)} & Whole-body &  \makecell{Admittance\\ controller} & No & No & Yes \\
\hline
\hline
\end{tabular}
\label{tab:related_work}
\end{table*}

Previous studies on controlling interaction forces with floating-base lower-limb exoskeletons with feet can be divided into three main categories.

\subsubsection{\textcolor{black}{Whole-body dynamics compensation with no interaction force feedback}}

A lower-limb exoskeleton can be modeled as a floating-base system considering all links and joints together, similar to a bipedal robot~\cite{Pratt1997, Mistry2020}. 

Vantilt et al.~\cite{VantiltJ2019Model-basedmovements} implemented a whole-body floating-base model while considering external contact constraints such as ground contact or contact with a stool. They estimated the contact forces based on a complementary energy method~\cite{Vantilt2018}. However, this method does not allow the calculation of interaction torques at the non-static contact points between the human and exoskeleton. Camardella et al.~\cite{Camardella2021} also modeled the full-body dynamics based on the user-dependent continuous interpolation between left and right stance dynamics.
These studies compensated for the gravitational and Coriolis torques calculated using the whole-body model. However, modeling errors were not corrected with interaction force feedback, and desired interaction torques were used only as a feedforward term. 

\subsubsection{\textcolor{black}{Simplified double pendulum model with interaction force feedback}}
Andrade et al.~\cite{Andrade2019, Andrade2021}, and Tue et al.~\cite{Tu2020} employ a simpler, non-whole-body model of the exoskeleton to help estimate interaction torques. In this simplified model, each leg is modeled as an independent double pendulum 2R robot. When the leg is in swing and stance, the hip and ankle joints are assumed to be pinned, respectively.
This model is used to calculate the feedforward compensation torques, which are subtracted from the joint torque measurements to estimate the interaction torques between the human and exoskeleton.
Interaction torque estimation is used as the input of a proportional-derivative (PD) or admittance controller.

Since the backpack is not modeled and the legs are modeled independently, the dynamic compensation torques and interaction torque estimates are not accurate for the stance leg. With this approach, the stance leg carries only its own weight instead of the weight of the entire exoskeleton. While this might be acceptable for light or grounded exoskeletons, it is not accurate for exoskeletons designed for overground walking that provide a high level of support and long battery life.

\subsubsection{\textcolor{black}{User-dependent or user-instrumented approaches}}
Since physical parameters and walking gaits vary from user to user, some interaction force controllers require additional sensors to be affixed to the user and/or custom tuning for each user. 

Sharifi et al.~\cite{Sharifi2021} implemented a neural network scheme to learn the whole-body dynamics of the exoskeleton together with the user. While this method does not require additional sensors to measure the joint torques, the main drawback is user-dependent learned dynamics, and a training phase is required for each user. Kazerooni et al.~\cite{Kazerooni2006} implemented a hybrid controller where the exoskeleton's stance leg follows the joint positions of the user to provide haptic transparency. The joint positions of the user's leg are measured by inclinometers attached to it. Moreover, the proposed method is for haptic transparency only; if the desired interaction force is non-zero, a model of the user-dependent interaction properties are needed.



\subsubsection{Summary}
Selected features of interaction force/torque control studies on floating-base lower-limb exoskeletons that contact the ground are presented in Table~\ref{tab:related_work}. These studies can be categorized into three main groups. First, there are studies where the whole-body dynamics of the exoskeleton are compensated with no interaction force feedback~\cite{Camardella2021, VantiltJ2019Model-basedmovements, Sharifi2021}. Second, some studies modeled the exoskeleton as two independent legs with no consideration of the backpack~\cite{Andrade2019, Andrade2021, Tu2020}. These studies implemented PD or admittance controllers based on the interaction force/torque error. Third, there are methods that require additional instrumentation on the user~\cite{Kazerooni2006} or user-dependent learning for model or gait state estimation~\cite{Sharifi2021, Camardella2021}. \textcolor{black}{None of these proposed control strategies implements optimization methods to enforce physical or safety limits to avoid actuator saturation or to constrain the motion to satisfy safety conditions.} 

To the best of our knowledge, there is no study that calculates interaction forces with full-body dynamics during the whole gait cycle and uses this estimation in a closed-loop control for a floating-base lower-limb exoskeleton that is in contact with the ground.

\subsection{Contributions}
In this paper, we propose a novel subject-independent method to calculate interaction forces during both single and double stance configurations based on a whole-body dynamic model. Moreover, we propose constrained optimization combined with a virtual model controller to achieve desired interaction torques while considering physical and safety constraints. \textcolor{black}{We call the proposed method \emph{whole-exoskeleton closed-loop compensation} (WECC) because whole-body dynamics of the exoskeleton is compensated and considered in the closed-loop controller.}

We evaluated the transparency and spring-damper haptic-rendering performance of our proposed WECC controller on treadmill walking. We compared the performance of the WECC controller to a control approach based on a simplified dynamic model of the exoskeleton (the ``simplified'' condition), where the legs are modeled as double pendulums independent of one another and the backpack is ignored~\cite{Andrade2021Role, Andrade2019, Andrade2021, Tu2020}. 
For transparency evaluations, we also tested the performance of a passive version of the exoskeleton with disassembled drives (the ``no-drive'' condition). 

We observed that the transparency of the swing leg was similar under the three conditions (WECC, simplified, and no-drive), but our proposed WECC method outperformed the simplified controller and the no-drive condition for the stance leg. In nonzero impedance haptic rendering, the WECC and simplified controllers performed similarly for the swing leg, but our WECC controller tracked the desired interaction torques better for the stance leg.

The core contributions of this article are as follows: 
\begin{enumerate}
	\item A continuous whole-body forward dynamical model for the whole gait cycle is developed based on the ratios of the vertical ground reaction forces at the feet.
	\item A novel method to estimate the interaction forces during both single and double stance configurations is proposed.
	\item A constrained-optimization-based virtual mass controller is implemented that considers the underactuation of the exoskeleton and its physical/safety constraints to control the interaction torques.
	\item The transparency and haptic rendering performance of our proposed WECC controller is compared to a state-of-the-art controller based on a simplified dynamic model.
\end{enumerate}

\section{System Description and Modeling }
\subsection{Lower-limb Exoskeleton: ExoMotus-X2}
\label{ssec:X2}

\begin{figure}[t!]
\centering
\includegraphics[width = 1\columnwidth]{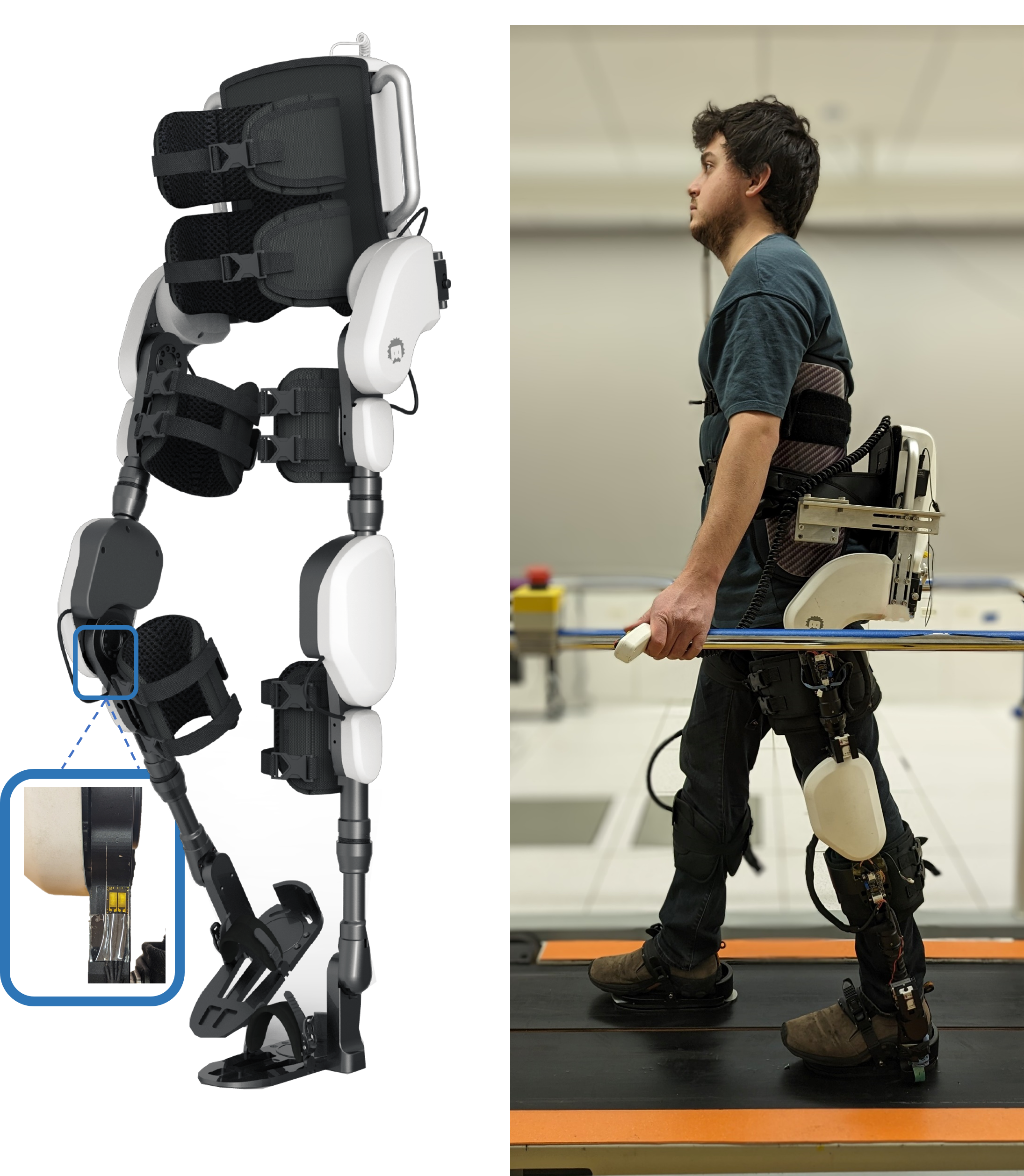}
\caption{ExoMotus-X2 lower-limb exoskeleton, \textcolor{black}{featuring a close-up view of the strain gauge implementation} (left) and a user wearing the exoskeleton (right).}
\label{fig:x2}
\end{figure}

A four-degrees-of-freedom (DoF) lower limb exoskeleton (ExoMotus-X2, Fourier Intelligence, Singapore), shown in Figure~\ref{fig:x2}, was adapted and used as a platform to test the proposed interaction torque controller. This exoskeleton has four total active degrees of freedom at the hip and knee joints, and passively allows motion at the ankle joints. Each active joint is driven by a Maxon EC60 100~W motor with a 1:122.5 reduction ratio obtained via harmonic gearing and a belt. Copley Accelnet ACK-055-06 motor drivers are used to control the motors. A peak torque of 80~Nm and a peak velocity of 3.2~rad/s can be obtained at each joint.

Due to the high friction caused by the large reduction ratio, it is not feasible to estimate the joint torques by only measuring the current through the motors. Therefore we modified the ExoMotus-X2 exoskeleton to add joint torque sensors at each driven joint. Each torque sensor consists of a full Wheatstone bridge with four strain gauges on the limb just distal to the joint, similar to the method presented by Claros et al.~\cite{Claros2016}. These custom joint torque sensors provide measurements with a resolution of 0.3~Nm and 0.1~Nm at the hip and knee joints, respectively. \textcolor{black}{The strain gauges were calibrated by immobilizing the joints and hanging known weights at known locations
while the exoskeleton was suspended by backpack attachment points. A close-up view of the strain gauge is shown in Figure~\ref{fig:x2}}

An inertial measurement unit (IMU) (Tech-IMU V4 by Technaid S.L.) is mounted on the backpack to measure the orientation and angular velocity of that link. Using the encoder measurements from each motor and orientation estimation of the backpack, the orientation of each link in the sagittal plane with respect to the gravity vector is estimated. In addition, eight 3-DoF force plates (9047B, Kistler) under a custom instrumented split-belt treadmill are used to measure ground reaction forces at each foot\footnote{\textcolor{black}{
The external footplates provide ground truth, but the controller uses only relative ground reaction forces rather than absolute ground reaction forces, and estimates of these relative forces may be provided by sensors on the exoskeleton itself. This is discussed further in Sections~\ref{sec:int_torque_control} and \ref{sec:Discussion}.}}. Furthermore, to have a tighter connection with the upper body, an additional trunk brace was adopted from another lower-limb exoskeleton (Exo-H3, Technaid S.L.).

Communication with motors and onboard sensors is established over a CAN bus using the CANOpen communication protocol. To minimize communication delay and allow real-time visualization and tuning capabilities, an external PC is connected to the exoskeleton by a tether. Analog force plate readings are acquired with an Arduino Mega board connected to the same PC. The controllers and applications are implemented on an open-source software stack called CANOpen Robot Controller (CORC)~\cite{FongCanOpenDevelopment} based on ROS and C++. High-level controller commands and sensor measurements are updated at 333~Hz, while the low-level current controller of the motors runs at 15~kHz.

\subsection{Dynamic Model}
\label{ssec:dynamic_model}
Our proposed WECC control method is based on a whole-body model where we incorporate the saggital plane dynamics of every link of the exoskeleton, including the backpack. 

\subsubsection{Whole-body dynamics}

\begin{figure}
\centering
\begin{tikzpicture}[->,>=stealth',shorten >=1pt,auto,node distance=3cm,
                    semithick]
  \tikzstyle{every state}=[draw=black, text=black, minimum size=1cm, align=center]

  \node[state] (ls)                    {Left\\stance};
  \node[state] (fly) [above right of=ls] {Flight};
  \node[state] (ds) [below right of=ls] {Double\\stance};
  \node[state] (rs) [below right of=fly] {Right\\stance};

  \path (ls) edge  [bend left]    node[sloped] {${F_{\text{l}}} < F_{\text{lim}}$} (fly)
             edge                 node[sloped] {${F_{\text{r}}} \geq F_{\text{lim}}$} (ds)
             edge  [loop left]    node[below,pos=0.2] {${F_{\text{l}}} \geq F_{\text{lim}}$} (ls)
        (fly) edge [loop above]   node[left,pos=0.2] {${F_{\text{l}}}, {F_{\text{r}}} < F_{\text{lim}}$} (fly)
              edge [bend left]    node[sloped] {${F_{\text{r}}} \geq F_{\text{lim}}$} (rs)
              edge                node[sloped, below] {${F_{\text{l}}} \geq F_{\text{lim}}$} (ls)
        (rs) edge   [bend left] node[sloped, below] {${F_{\text{l}}} \geq F_{\text{lim}}$} (ds)
             edge   node[sloped, below] {${F_{\text{r}}} < F_{\text{lim}}$} (fly)
             edge [loop right] node[above,pos=0.2] {${F_{\text{r}}} \geq F_{\text{lim}}$} (rs)
        (ds) edge [loop below] node[right,pos=0.2] {${F_{\text{l}}}, {F_{\text{r}}} \geq F_{\text{lim}}$} (ds)
            edge  [bend left]  node[sloped, below] {${F_{\text{r}}} < F_{\text{lim}}$} (ls)
            edge   node[sloped] {${F_{\text{l}}} < F_{\text{lim}}$} (rs);
\end{tikzpicture}
    \caption{\textcolor{black}{State machine structure of the gait states.} The variables ${F_{\text{l}}}$ and ${F_{\text{r}}}$ are the left foot and right foot vertical force readings, respectively. The threshold force that triggers state change is represented by $F_{\text{lim}}$. \textcolor{black}{ This threshold can be an absolute force value or a ratio of the total vertical forces.} \textcolor{black}{When this state machine is used in real-time control}, states are not allowed to change more than once every 150~ms for robustness to noise. \textcolor{black}{During walking, the focus of this paper, the flight state never occurs.}}
    \label{fig:gait_state}
\end{figure}
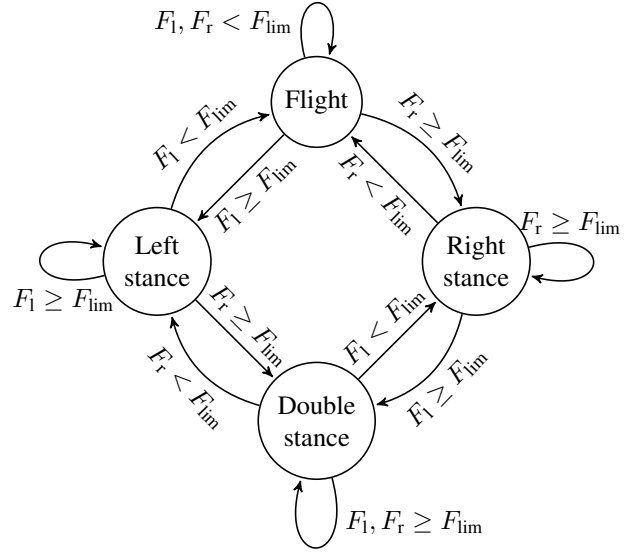

The exoskeleton is modeled as a floating-base five-link mechanism with four active joints. The dynamics of the human user are not modeled but are incorporated into the exoskeleton dynamics as a source of external torque at each joint. For each gait state, the constraints on the system and the equations of motion are modified accordingly. \textcolor{black}{These gait states are no ground contact (\emph{flight}), contact by a single foot (\emph{single stance}), and contact by two feet (\emph{double stance}).} 
Figure~\ref{fig:gait_state} presents the gait state machine. 

\paragraph{Single stance}

\begin{figure}
\centering
\begin{subfigure}{0.24\textwidth}
\centering
    \includegraphics[height=\textwidth]{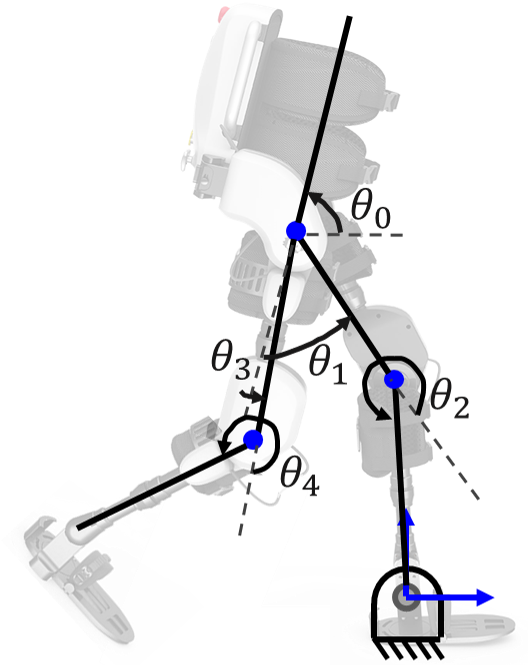}
    \caption{Single stance}
    \label{fig:ss}
\end{subfigure}
\begin{subfigure}{0.24\textwidth}
\centering
    \includegraphics[height=\textwidth]{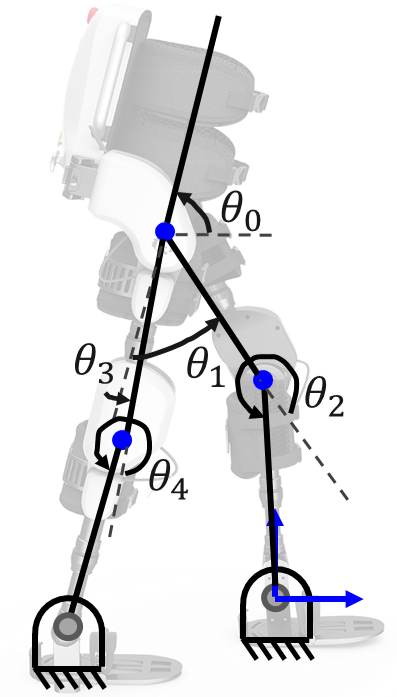}
    \caption{Double stance}
    \label{fig:double_stance}
\end{subfigure}%
\\
\begin{subfigure}{0.24\textwidth}
\centering
    \includegraphics[height=\textwidth]{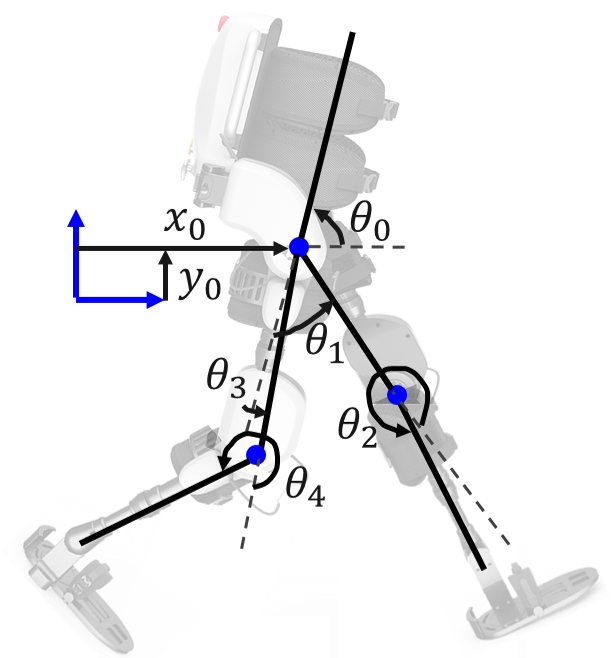}
    \caption{Flight}
    \label{fig:flight}
\end{subfigure}%
\begin{subfigure}{0.24\textwidth}
\centering
    \includegraphics[height=\textwidth]{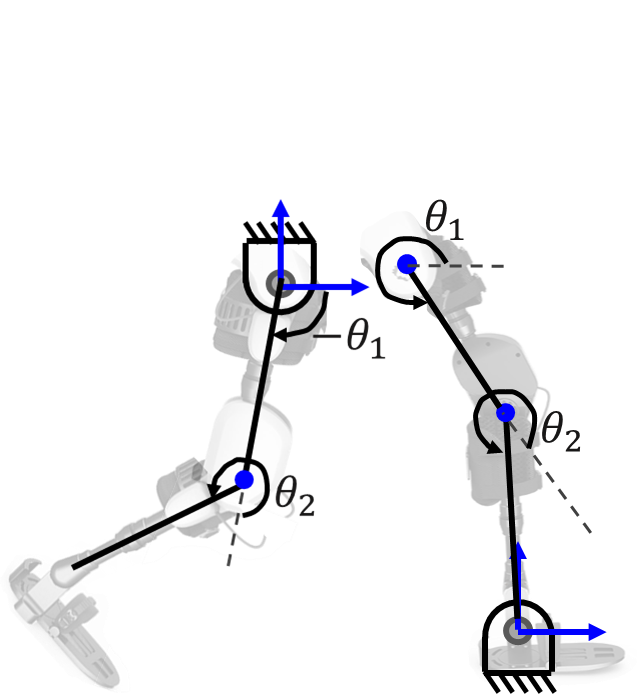}
    \caption{Simplified}
    \label{fig:simplfied}
\end{subfigure}
        
\caption{Different gait states and representation of the generalized coordinates. Whole-body model representations are shown in sub-figures (a), (b), and (c). Sub-figure (d) corresponds to the stance and swing phases of the simplified model.}
\label{fig:state_figures}
\end{figure}

In single stance, the ankle of the stance leg is assumed to be hinged to the ground. With this assumption, the exoskeleton has five degrees of freedom. The generalized coordinates and equations of motion are 
\begin{equation}
    \mt{M}_i(\mt{q})\ddot{\mt{q}} + \mt{b}_i(\mt{q},\dot{\mt{q}}) + \mt{g}_i(\mt{q}) = \mt{S}^\top\mt{\tau}_\text{joint} + \mt{\tau}_\text{int}, \;\: i\in \{\text{ls}, \text{rs}\},
    \label{eq:SSEoM}
\end{equation}
where $\mt{q} = [\theta_0, \theta_1, \theta_2, \theta_3, \theta_4]^\top$ are the generalized coordinates corresponding to the backpack, hip, and knee angles, as shown in Figure~\ref{fig:ss}. The matrix and vectors $\mt{M}_i \in \mathbb{R}^{5\times 5}, \mt{b}_i \in \mathbb{R}^5, \mt{g}_i \in \mathbb{R}^5$, are the mass matrix,  Coriolis and centrifugal torques, and gravitational torques respectively during left stance ($i = \text{ls}$) and right stance ($i = \text{rs}$). The vector $\mt{\tau}_\text{joint} \in \mathbb{R}^4$ corresponds to joint torques, and $\mt{S} = [0_{4\times 1}, \mathbb{I}_{4\times 4}]$ is a selection matrix of actuated joints. The interaction torques applied to the exoskeleton by the user are given by the vector $\mt{\tau}_\text{int} \in \mathbb{R}^5$.


\paragraph{Double stance}

\textcolor{black}{This state can be expressed as an extension of the single stance model where one ankle is assumed to be hinged on the ground and external forces are applied on the other ankle,}
\begin{equation}
\begin{split}
    &\mt{M}_i(\mt{q})\ddot{\mt{q}} + \mt{b}_i(\mt{q},\dot{\mt{q}}) + \mt{g}_i(\mt{q}) = \mt{S}^\top\mt{\tau}_\text{joint} + \mt{\tau}_\text{int} + \mt{J}_j^\top\boldsymbol{\Gamma}_j, \\ &(i,j)\in \{(\text{ls}, \text{r}), (\text{rs}, \text{l})\},
    \label{eq:DS1}
\end{split}
\end{equation}
where $J_j$ and $\Gamma_j$ corresponds to the contact Jacobian of the left ($j$ = l) or right ($j$ = r) ankle and the external forces at that point, respectively. Multiplying Equation~\eqref{eq:DS1} by the transpose of the constraint matrix $\mt{H} \in \mathbb{R}^{5\times 3}$, where $\mt{J}\mt{H} = 0$, results in the following equation of motion:
\begin{equation}
\begin{split}
    &\mt{H}^\top(\mt{M}_i(\mt{q})\ddot{\mt{q}} + \mt{b}_i(\mt{q},\dot{\mt{q}}) + \mt{g}_i(\mt{q}) - \mt{\tau}_\text{int}) = \mt{H}^\top\mt{S}^\top\mt{\tau}_\text{joint}, \\ &i\in \{\text{ls}, \text{rs}\}.
    \label{eq:DS2}
\end{split}
\end{equation}
\textcolor{black}{The two position constraints on the other ankle represented in the constraint matrix $\mt{H}$ map the five-dimensional Equation~\eqref{eq:DS1} onto the three-dimensional controllable space. Any three of the five coordinates may be chosen as the generalized coordinates.\footnote{Interested readers may refer to~\cite{Li2021} for the details of the constraint matrix $\mt{H}$ and Equations~\eqref{eq:DS1}--\eqref{eq:DS2}.} Since there are only three DoF, the system is overactuated during double stance; there exists a one-dimensional null space of internal ``cocontraction'' torques that have no impact on motion.
Overactuation in double stance complicates the estimation of interaction torques, as discussed in Section~\ref{sec:int_torque_control}, and may cause a discontinuity in the equations of motion and the dynamic compensation torques during transitions between single and double stance.}


To sidestep these issues, we approximate double stance as a transition from left stance to right stance or vice versa. 
\begin{equation}
    \mt{M}_\text{ds}(\mt{q})\ddot{\mt{q}} + \mt{b}_\text{ds}(\mt{q},\dot{\mt{q}}) + \mt{g}_\text{ds}(\mt{q}) = \mt{S}^\top\mt{\tau}_\text{joint} + \mt{\tau}_\text{int} ,
    \label{eq:DSEoM}
\end{equation}
where
\begin{align}
        \mt{M}_\text{ds} &= \alpha\mt{M}_\text{ls} + (1-\alpha)\mt{M}_\text{rs},
    \label{eq:DSEoM1} \\
    \mt{b}_\text{ds} &= \alpha\mt{b}_\text{ls} + (1-\alpha)\mt{b}_\text{rs},
    \label{eq:DSEoM2}     \\
    \mt{g}_\text{ds} &= \alpha\mt{g}_\text{ls} + (1-\alpha)\mt{g}_\text{rs}.
    \label{eq:DSEoM3} 
\end{align}
The variable $\alpha$ in Equations~\eqref{eq:DSEoM1}--\eqref{eq:DSEoM3} corresponds to the interpolation factor from left-stance dynamics to right-stance dynamics. \textcolor{black}{These interpolated dynamics are particular solutions to the equations of motion~\eqref{eq:DS2}, as proven in Appendix~\ref{app:ds}.} The interpolation factor $\alpha$  is calculated based on the ratio of the left vertical ground reaction force to the sum of both vertical ground reaction forces
\begin{equation}
    \alpha = \frac{{F_{\text{l},y}}}{{F_{\text{l},y}} + {F_{\text{r},y}}},
    \label{eq:interpolationFactor}
\end{equation}
where ${F_{\text{l},y}}$ and ${F_{\text{r},y}}$ are left and right vertical ground reaction forces, respectively. \textcolor{black}{The rationale for defining the $\alpha$ based on the ratios of the vertical ground reaction forces is presented in Appendix~\ref{app:ds}.} \textcolor{black}{Interpolating the equation of motion from left-stance dynamics to right-stance dynamics allows smooth continuous modeling of the whole gait cycle that is consistent with the weight transition of the exoskeleton during walking.} The difference of this method from the work of Camardella et al.~\cite{Camardella2021} is that their kinematic-based gait segmentation is user-dependent and requires a training procedure.

\paragraph{\textcolor{black}{Flight}}
In the flight state, the only external forces on the exoskeleton are due to interaction with the user. \textcolor{black}{While this state does not occur during walking, we use this state's equation of motion for exoskeleton parameter estimation, as described in Section~\ref{ssec:param_est}.} The equation of motion in generalized coordinates is
\begin{equation}
    \mt{M}_\text{fly}(\mt{q}_\text{fly})\ddot{\mt{q}}_\text{fly} + \mt{b}_\text{fly}(\mt{q}_\text{fly},\dot{\mt{q}}_\text{fly}) + \mt{g}_\text{fly}(\mt{q}_\text{fly}) = \boldsymbol{S}_\text{fly}^\top\mt{\tau}_\text{joint} + \mt{\tau}_\text{int}^\text{fly},
    \label{eq:FlyingEoM}
\end{equation}
where $\mt{q_\text{fly}} = [x_0, y_0, \theta_0, \theta_1, \theta_2, \theta_3, \theta_4]$ is a vector of generalized coordinates corresponding to the linear and angular positions of the backpack and the hip and knee joint angles, as shown in Figure~\ref{fig:flight}. The variable $\mt{M}_\text{fly} \in \mathbb{R}^{7\times 7}$ is the mass matrix, $\mt{b}_\text{fly} \in \mathbb{R}^7$ is a vector of Coriolis and centrifugal forces, and $\mt{g}_\text{fly} \in \mathbb{R}^7$ a vector of gravitational forces. The variable $\boldsymbol{S}_\text{fly} = [0_{4\times 3}, \mathbb{I}_{4\times 4}]$ is a selection matrix of actuated joints, and $\mt{\tau}_\text{int}^\text{fly} \in \mathbb{R}^7$ is a vector of interaction forces and torques applied to the exoskeleton by the user.

\subsubsection{Simplified dynamics}

\textcolor{black}{The dynamics described above are used in our proposed WECC controller. A simpler model, described here, considers each leg of the exoskeleton as an independent double pendulum and excludes the backpack}~\cite{Hwang2015, Andrade2021Role, Mayag2022, vanDijkW2013Improvingwalking, Tu2020, ZanottoD2013Improvingcuffs}. We use this model for comparison to our approach. The hip joint is considered grounded for a swing leg, and the ankle joint is considered fixed to the ground for a stance leg, as shown in Figure~\ref{fig:simplfied}.

The equation of motion can be written 
\begin{equation}
    \mt{M}_i(\mt{q})\ddot{\mt{q}} + \mt{b}_i(\mt{q},\dot{\mt{q}}) + \mt{g}_i(\mt{q}) = \mt{\tau}_\text{joint} + \mt{\tau}_\text{int},
    \;\: i\in \{\text{st}, \text{sw}\},
    \label{eq:simpleEoM}
\end{equation}
where $\mt{q} = [\theta_1, \theta_2]^\top$ are the hip and knee joint positions, respectively. The variables $\mt{M}_i \in \mathbb{R}^{2\times 2}, \mt{b}_i \in \mathbb{R}^2, \mt{g}_i \in \mathbb{R}^2$, are the mass matrix,  Coriolis and centrifugal torques, and gravitational torques, respectively, during stance ($i = \text{st}$) and swing ($i = \text{sw}$) \textcolor{black}{of a single leg}.

\subsection{Parameter Estimation}
\label{ssec:param_est}
Inertial parameters such as mass, mass moment of inertia, and center of mass (CoM) locations of each link were estimated from the CAD model of the ExoMotus-X2 exoskeleton. \textcolor{black}{Dynamic terms such as $\mt{M}$, $\mt{b}$, and $\mt{g}$ are defined as a function of these parameters, adjustable link lengths, and instantaneous joint states. Their values are calculated at every time instant.} Some of these parameters are presented in Table~\ref{tab:x2_params}. Note that the largest inertias come from the apparent rotor inertia \textcolor{black}{which is proportional to the square of the reduction ratio (RR = 122.5).}

\begin{table}
\centering
\caption{Estimated masses and inertias (about axes through the center of mass and perpendicular to the saggital plane) of links and motor rotors of the ExoMotus-X2 exoskeleton.}
\begin{tabular}{c| c c} 
\cline{1-3}
\rule{0pt}{2ex}  
 Link & $m$ [kg] & $L$ [kgm\textsuperscript{2}] \\ \cline{1-3}
  Backpack & 10.3 & 0.201 \\ 
  Upper thigh & 0.7 & 0.002197 \\ 
  Lower thigh & 2.14 & 0.005901 \\
  Upper shank & 0.64 & 0.002346 \\
  Lower shank & 0.47 & 0.0007941 \\
  Foot & 1.25 & 0.004441 \\
  Rotor (apparent) & - & $1.246 10^{-4} \times 122.5^2 = 1.87$  \\ \cline{1-3}
\end{tabular}
\label{tab:x2_params}
\end{table}

The high gear ratios also result in significant friction at each joint. Friction at the joints is modeled as viscous friction combined with Coulomb friction as a function of joint speed,
\begin{equation}
    \mt{\tau}_\text{friction} = \mt{c_0}\odot\sign{\dot{\mt{\theta}}} + \mt{c_1}\odot\dot{\mt{\theta}},
    \label{eq:friction_model}
\end{equation}
\textcolor{black}{where $\dot{\mt{\theta}}$ is the vector of joint velocities and $\mt{c_0}$ and $\mt{c_1}$ are the vectors of Coulomb and viscous coefficients, respectively. The symbol $\odot$ indicates element-wise Hadamard product.}


\textcolor{black}{To conduct the parameter estimation experiments, the exoskeleton was suspended by backpack attachment points. Chirp motor torque commands with increasing frequency up to $3$~Hz at different amplitudes were simultaneously given to the hip and knee joints of a single leg, 
\begin{equation}
    \tau_\text{motor} = A_{i}\sin(6\pi t^2/T)
    \label{eq:estimation_func},
\end{equation}
where $A_i \in \{7.5, 10, 12.5, 15\}$~Nm and $T=60$~s for each amplitude. 
Joint positions were measured during the experiments.}


\textcolor{black}{The joint torques, $\mt{\tau}_\text{joint}$, in Equations~\eqref{eq:SSEoM}, \eqref{eq:DS1} and \eqref{eq:FlyingEoM} are the sum of motor torques, $\mt{\tau}_\text{motor}$, and friction compensation torques, $\mt{\tau}_\text{friction}$,
\begin{equation}
    \mt{\tau}_\text{joint} = \mt{\tau}_\text{motor} + \mt{\tau}_\text{friction}
    \label{eq:joint_torque}.
\end{equation}
}

As there was no contact with the ground at this configuration, the equation of motion of the flight state was used to estimate the frictional parameters. Equation~\eqref{eq:FlyingEoM} is rewritten to include viscous and Coulomb friction constants \textcolor{black}{by substituting $\mt{\tau}_\text{joint}$ in Equation~\eqref{eq:joint_torque} into Equation~\eqref{eq:FlyingEoM},}

\begin{equation}
\begin{split}
    \mt{M}_\text{fly}(\mt{q}_\text{fly})\ddot{\mt{q}}_\text{fly} + \mt{b}_\text{fly}(\mt{q}_\text{fly},\dot{\mt{q}}_\text{fly}) + \mt{g}_\text{fly}(\mt{q}_\text{fly}) = \\ \mt{S}_{\text{fly}}^\top(\mt{\tau}_\text{motor} + \mt{c_0}\odot\sign{\dot{\mt{\theta}}} + \mt{c_1}\odot\dot{\mt{\theta}})  +  \mt{\tau}_\text{int}.
    \label{eq:FlyingEoMFriction}
\end{split}
\end{equation}

Recorded backpack and joint angular positions, joint angular velocities, joint angular accelerations, and CAD-based inertial parameters were used to calculate $\mt{M}_\text{fly}$, $\mt{b}_\text{fly}$, and $\mt{g}_\text{fly}$ at each time instant. \textcolor{black}{Joint accelerations were calculated by taking the derivative of the joint velocities numerically by the nine-point difference method~\cite{BARKLEYROSSER1975351}. Then, in Matlab, accelerations were low-pass filtered with a cutoff frequency of 10~Hz, while also incorporating delay compensation.} Since the exoskeleton was hung from the backpack at a fixed location, $\ddot{x}$ and $\ddot{y}$ were negligible. Also, as there was no user in the exoskeleton, rotational components of $\mt{\tau}_\text{int}$ were zero. The coefficients $\mt{c_0}$ and $\mt{c_1}$ were solved using \textcolor{black}{the recorded data by least-square minimization on Equation~\eqref{eq:FlyingEoMFriction}.} Estimated friction parameters for each joint and the $R^2$ value of the fit are presented in Table~\ref{tab:friction_params}.

\begin{table}
\centering
\caption{Estimated friction parameters of the ExoMotus-X2 exoskeleton and R\textsuperscript{2} of the fit.}
\begin{tabular}{c| c c c} 
\cline{1-4}
\rule{0pt}{2ex}  
 Joint & $c_0$ [Nm] & $c_1$[Nms] & R\textsuperscript{2} \\ \cline{1-4} 
  Left hip & 5.01 & 4.58 & 0.96 \\ 
  Right hip & 3.26 & 4.67 & 0.94 \\
  Left knee & 4.30 & 3.25 & 0.96 \\
  Right knee & 4.45 & 5.16  & 0.96 \\ \cline{1-4}
\end{tabular}
\label{tab:friction_params}
\end{table}
\section{Interaction Torque Controller}
\label{sec:int_torque_control}

\subsection{Interaction Torque Estimation}

Estimating the interaction torques between the human and exoskeleton plays a significant role in the transparency and haptic rendering capabilities of the robot. Placing force/torque sensors at every contact point between the human and exoskeleton would theoretically be ideal for measuring the interaction.
However, due to the \textcolor{black}{distributed contacts between the human and the exoskeleton and} practical challenges of this approach, we chose to estimate the interaction torques through joint torque measurements. Joint torques were measured using our custom joint-torque sensors mentioned in Section~\ref{ssec:X2}.


Because the ExoMotus-X2 exoskeleton was designed for individuals with sensorimotor impairments and therefore limited walking ability, we estimated the interaction torque during single and double stance only, excluding the flight state.

\subsubsection{Single stance interaction torque estimation}

We estimate the interaction torque during single stance by subtracting the gravitational and Coriolis forces from the joint torque readings:
\begin{equation}
    \mt{\tau}_\text{int} = -\mt{S}^\top\mt{\tau}_\text{joint} + \mt{b}_i(\mt{q},\dot{\mt{q}}) + \mt{g}_i(\mt{q}),  \;\: i\in \{\text{ls}, \text{rs}\}.
    \label{eq:SS_interaction_estimation}
\end{equation}
\textcolor{black}{As shown in Table~\ref{tab:x2_params}, the largest contributors to the inertial terms are the apparent rotor inertias due to the high reduction ratios. However, since the strain gauges are placed at the link (i.e., after the motor), the torque measurements are not affected by the apparent rotor inertia. In addition, because the links have relatively low inertia and the walking speed during the experiment is not high, the effects of inertial forces are significantly smaller compared to gravitational forces. Therefore, acceleration-dependent inertial components are neglected in the interaction torque estimation.}


It is worth noting that $\mt{\tau}_\text{int} \in \mathbb{R}^5$, which includes the interaction torque at the backpack joint in addition to the four leg joints.

\subsubsection{Double stance interaction torque estimation}

\textcolor{black}{Interaction torque estimation can be expressed by extending the left and right stance model with the additional external contact force at the right and left foot, respectively,}
\begin{align}
        \mt{\tau}_\text{int} &= -\mt{S}^\top\mt{\tau}_\text{joint} + \mt{b}_{\text{ls}}(\mt{q},\dot{\mt{q}}) + \mt{g}_{\text{ls}}(\mt{q}) - J_{\text{r}}^\top \mt{F_{\text{r}}}
    \label{eq:DS_interaction_estimation_1} \\
    \mt{\tau}_\text{int} &= -\mt{S}^\top\mt{\tau}_\text{joint} + \mt{b}_{\text{rs}}(\mt{q},\dot{\mt{q}}) + \mt{g}_{\text{rs}}(\mt{q})  - J_{\text{l}}^\top \mt{F_{\text{l}}},
    \label{eq:DS_interaction_estimation_2}
\end{align}
\textcolor{black}{where $\mt{J}_i$ is the ankle Jacobian and $F_i$ is the external force applied on the left ($i = \text{l}$) and right ($i = \text{r}$) foot.}
Combining Equations~\ref{eq:DS_interaction_estimation_1} and~\ref{eq:DS_interaction_estimation_2} results in ten equations with rank seven. Nine unknowns in these equations are the five-dimensional interaction torque vector, $\mt{\tau}_\text{int}$, and the two-dimensional contact forces at the left ($\mt{F_{\text{l}}}$) and right ($\mt{F_{\text{r}}}$) feet. Since there are seven independent equations and nine unknowns, it is not possible to uniquely solve for the interaction torque vector without additional information.

\textcolor{black}{
Ratios of the vertical ground reaction forces (GRF) and the direction of the GRF in the sagittal plane are used to resolve the redundancy. In this study, we used external force plates to measure these terms, but they can also be obtained by wearable sensors, or sensors instrumented onto the exoskeleton, as explained in Section~\ref{ssec:overground_walking}}.


\textcolor{black}{The vertical ground reaction force ratio $\alpha$ is introduced in Equation~\eqref{eq:interpolationFactor} and is also used for the equations of motion during double stance. The direction of the force can be expressed as the ratio of the horizontal and vertical forces,
\begin{equation}
 \gamma = \frac{F_{\text{l,x}}}{F_{\text{l,y}}}.
 \label{eq:gamma}
\end{equation}
}

\textcolor{black}{Equations~\eqref{eq:interpolationFactor} and \eqref{eq:DS_interaction_estimation_1}-\eqref{eq:gamma} are converted to a system of linear equations as $\textbf{K} \boldsymbol{\xi} = \textbf{m}$, where
\begin{equation}
\begin{split}
\textbf{K} &=
\begin{bmatrix}
    \mt{J}_{\text{r}}^\top & 0_{5 \times 2} & \mathbb{I}_{5\times5} \\
     0_{5 \times 2} & \mt{J}_{\text{l}}^\top & \mathbb{I}_{5\times5} \\
     [0, \; 1] & [0, \; -\alpha/(1-\alpha)] & 0_{1 \times 5} \\
     [0, \; 0] & [1, \;  -\gamma] & 0_{1 \times 5}
\end{bmatrix}, \\
\boldsymbol{\xi} &= 
\begin{bmatrix}
F_{\text{r,x}} & F_{\text{r,y}} & F_{\text{l,x}} & F_{\text{l,y}} & \mt{\tau}_\text{int}^\top
\end{bmatrix}^\top, \\
\textbf{m} &=
\begin{bmatrix}
-\mt{S}^\top\mt{\tau}_\text{joint} + \mt{b}_{\text{ls}}(\mt{q},\dot{\mt{q}}) + \mt{g}_{\text{ls}}(\mt{q}) \\
-\mt{S}^\top\mt{\tau}_\text{joint} + \mt{b}_{\text{rs}}(\mt{q},\dot{\mt{q}}) + \mt{g}_{\text{rs}}(\mt{q}) \\
0 \\
0
\end{bmatrix},
 \label{eq:ds_system_eq}
\end{split}
\end{equation}
}
\textcolor{black}{and solved for the interaction torques.} 

\subsubsection{Simplified interaction torque estimation}

With the simplified double pendulum model, interaction torques at the hip and knee joints can be estimated by subtracting the dynamic effects from the joint torque measurements,
\begin{equation}
    \mt{\tau}_\text{int} = \mt{\tau}_\text{joint} + \mt{b}_i(\mt{q},\dot{\mt{q}}) + \mt{g}_i(\mt{q}),  \;\: i\in \{\text{st}, \text{sw}\}.
    \label{eq:simplified_interaction_estimation}
\end{equation}
It is important to note that with the simplified method, $\mt{\tau}_\text{int} \in \mathbb{R}^{2}$ does not include the interaction torque at the backpack. Moreover, because the backpack link is not incorporated into the dynamic model, the backpack's weight is not subtracted, resulting in inaccurate interaction torque estimation for exoskeletons with heavy backpacks.

\subsection{Control}
\subsubsection{Virtual mass controller}
\begin{figure*}
\centering
\begin{tikzpicture}[->,>=stealth',shorten >=1pt,auto,node distance=3cm,
                    semithick]
                    
  \tikzstyle{block}=[rectangle, draw=black, text=black, minimum size=1cm, align=center]
  \tikzstyle{node}=[circle, draw=black, text=black, align=center]
  \tikzstyle{empty}=[text=black, align=center]

  \node[block] (MV)    {$\mt{M}_\text{virt}^{-1}$};
  \node[node] (refCircle)  [left = 1 cm of MV]  {-};
  \node[block] (HO) [right = 1.5 cm of MV] {
  $\displaystyle \argmin_{\textbf{x}} \lVert f(\textbf{x})\rVert = (\ddot{\mt{q}}, \mt{\tau}_{\text{motor}})$\\ s.t. \\ 
  $\textbf{A} \textbf{x} = \textbf{b} $ \\
  $ \textbf{D} \textbf{x}  \leq \textbf{f} $
  };
  \node[empty] (HO_empty) [above = -0.05 cm of HO] {Constrained Optimization};
  \node[block] (HR) [right = 2 cm of HO] {
  \includegraphics[width=.1\textwidth]{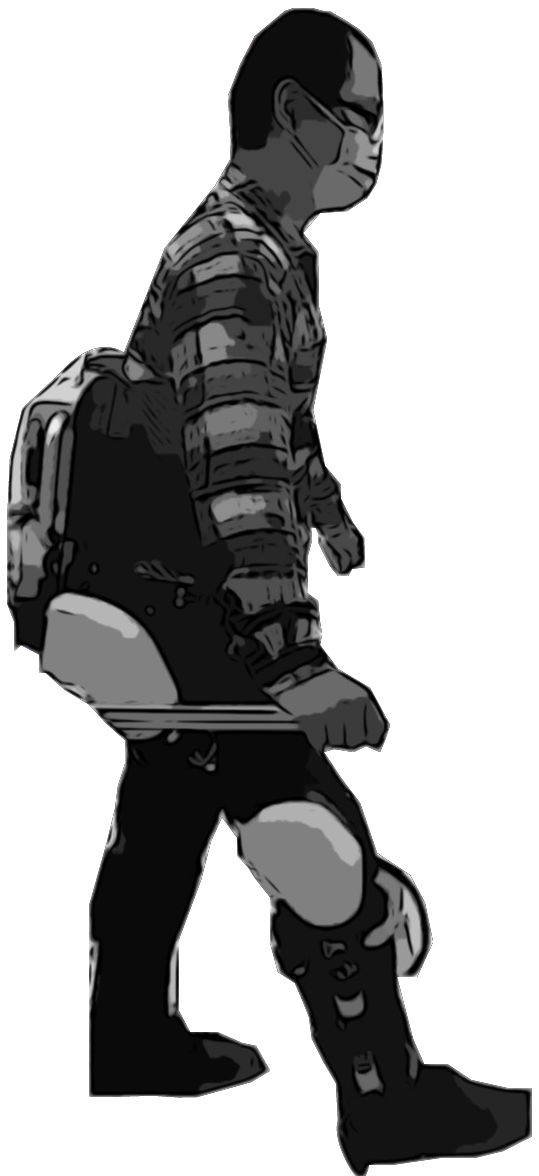}
  };
 \node[empty] (HR_empty) [above = -0.05 cm of HR] {Coupled Human-Robot};
  \node[empty] (HR_right) [right = 0.5 cm of HR] {};
  \node[block] (GS) [below = 0.75 cm of HO] {
  \begin{tikzpicture}[->,>=stealth',shorten >=1pt,auto,node distance=3cm,
                    semithick]
  \tikzstyle{every state}=[draw=black, text=black, minimum size=1cm, align=center]

  \node[state] (ls)                    {LS\\ $\alpha = 1$};
  \node[state] (ds) [below right = 0.5 cm of ls] {DS\\ $\alpha \in [0 \ 1]$};
  \node[state] (rs) [above right = 0.5 cm of ds] {RS\\ $\alpha = 0$};

  \path (ls) 
             edge                 node[sloped] {} (ds)
        (rs) 
             edge   [bend left] node[sloped, below] {} (ds)
        (ds) 
            edge  [bend left]  node[sloped, below] {} (ls)
            edge   node[sloped] {} (rs);
\end{tikzpicture}
  };
  \node[empty] (GS_right) [right = 4.3 cm of GS] {};
  \node[block] (FE) [below = 0.5 cm of GS] {Interaction Torque Estimator};
  \node[empty] (FE_right) [right = 4.63 cm of FE] {};
  \node[empty] (FE_left) [left = 3.53 cm of FE] {};
  \node[empty] (tau_des) [left = 1 cm of refCircle] {};
  
  \path (MV) edge node[] {$\ddot{\mt{q}}^*$} (HO);
  \path (HO) edge node[] {$(\ddot{\mt{q}}, \mt{\tau}_{\text{motor}})$} (HR);
  \path (GS) edge node[] {$\alpha$} (HO);
  \path (GS) edge node[] {$\alpha$} (FE);
  \draw [-] (HR.east) -| node[pos=0.5] {} (GS_right.center);
  \draw [->] (GS_right.center) -- node[above, pos=0.5] {$\mt{F}^\text{grf}$} (GS);
  \draw [->] (GS.west) -| node[near end] {$\alpha$} (MV.south);
   \draw [-] (GS_right.center) -- node[pos=0.5] {} (FE_right.center);
   \draw [->] (FE_right.center) -- node[above, pos=0.5] {$[\mt{q},\dot{\mt{q}}, \mt{\tau}_\text{joint}, \mt{F}^\text{grf}]$} (FE);
   \draw [->] (FE.west) -| node[pos = 0.9] {$\mt{\tau}_\text{int}$} (refCircle.south);
   \draw [->] (refCircle) -- node[pos=0.5] {$\mt{\tau}_\text{int}^{\text{err}}$} (MV);
   \draw [->] (tau_des) -- node[pos=0.5] {$\mt{\tau}_\text{int}^{\text{*}}$} (refCircle);
\end{tikzpicture}
    \caption{Schematic of the interaction force controller.}
    \label{fig:control_diagram}
\end{figure*}
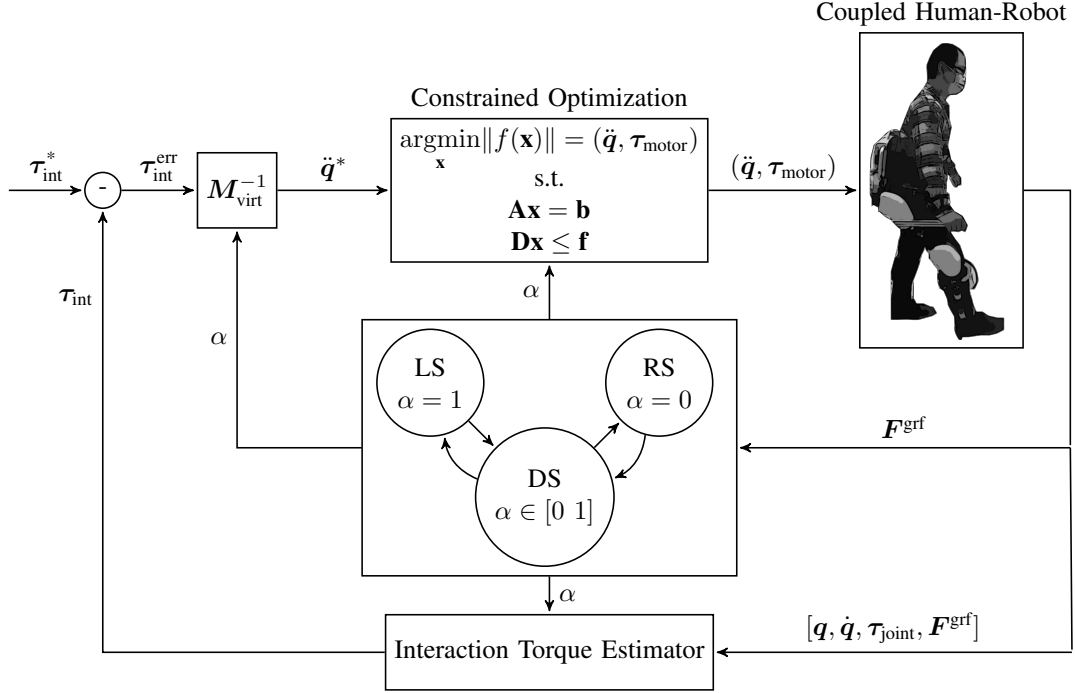

To control the interaction torques between the human and the exoskeleton, we implemented a \emph{virtual mass controller}~\cite{Fumagalli2013, Zimmermann2020} to simulate a desired mass or inertia, with no damping, of each generalized coordinate of the exoskeleton. This can be achieved by setting the desired generalized acceleration to
\begin{equation}
    \ddot{\mt{q}}^* = \mt{M}_\text{virt}^{-1}(\mt{\tau}_\text{int} - \mt{\tau}_\text{int}^*),
    \label{eq:desired_acc}
\end{equation}
where $\mt{\tau}_\text{int}^*$ is the desired interaction torque, e.g., zero for transparency or nonzero for other desired impedances.

Under the assumptions of perfect control of the generalized accelerations and accurate interaction torque estimation with little delay, this equation shows that the interaction torque error is proportional to the chosen virtual mass matrix $\mt{M}_\text{virt}$ and joint accelerations. Therefore, lowering $\mt{M}_\text{virt}$ results in better interaction torque tracking. Obviously, $\mt{M}_\text{virt}$ cannot be decreased below a certain limit due to the actuation limits, modeling errors, and communication delays. \textcolor{black}{Moreover, the limit on the virtual mass also depends on the interaction properties between the human and exoskeleton. While the damping due to the soft tissues of the user makes the system more stable, a very stiff connection would increase the minimum allowable virtual mass for a stable response.} 

We used a diagonal $\mt{M}_\text{virt}$ and tuned the virtual mass parameter for each generalized coordinate independently. \textcolor{black}{
The tuning process was done on a pilot user (a non-participant) by reducing each virtual mass as much as possible before compromising the stability of the control. The parameters were then verified with two other pilot users.} This process resulted in different virtual mass parameters for single stance and double stance, as shown in Table~\ref{tab:vm_values}. The virtual inertia of the backpack is relatively large to allow better control performance at the leg segments.

\begin{table}
\begin{center}
\caption{Virtual mass values for each link used during the experimentation.}\vspace{1ex}
    \label{tab:vm_values}
\begin{tabular}{cc|c}
\cline{1-3}
\rule{0pt}{2ex}  
Link & Gait State & Virtual Mass [kgm\textsuperscript{2}]    \\ \cline{1-3}
Backpack         & Single Stance & 2      \\
Backpack         & Double Stance & 3.5      \\
Thigh         & Single Stance & 0.7      \\
Thigh         & Double Stance & 1.2      \\
Shank         & Single Stance & 0.5      \\
Shank        & Double Stance & 0.87     \\ \cline{1-3}
\end{tabular}
\end{center}
\end{table}





\subsubsection{Constrained optimization}

While the desired acceleration $\ddot{\mt{q}}^*$ in Equation~\eqref{eq:desired_acc} is a five-dimensional vector, the exoskeleton has only four actuated joints. Therefore, it is not possible to perfectly track the desired generalized accelerations calculated by the virtual mass controller. Moreover, the maximum torque, power, and velocity of the motors bring additional limitations. To track the desired generalized acceleration as accurately as possible under physical and safety constraints, we use real-time constrained optimization with OSQP~\cite{osqp} to solve for $\textbf{x} = [\ddot{\mt{q}}^\top, \mt{\tau}_{\text{motor}}^\top]^\top \in \mathbb{R}^9$, where $\mt{\tau}_{\text{motor}}$ is a vector of torque \textcolor{black}{commands sent to the motor drivers}:
\begin{equation}
\begin{split}
\min_\textbf{x} \qquad &\lVert f(\textbf{x})\rVert \\
s.t. \qquad &\textbf{A} \textbf{x} = \textbf{b} \\
& \textbf{D} \textbf{x}  \leq \textbf{f}.
\end{split}
\label{eq:const_opt}
\end{equation}
The objective function $\lVert f(\textbf{x}) \rVert$ is the 2-norm of the difference between the actual and desired generalized accelerations,
\begin{equation}
f(\textbf{x}) =  [\mathbb{I}_{5\times5},\; 0_{5\times4}]\textbf{x} - \ddot{\mt{q}}^*.
\label{eq:obj_function}
\end{equation}




The equality constraint ensures that optimized variables satisfy the equation of motion of the corresponding gait state under physical limits,
\begin{equation} \label{eq:task_EOM}
\begin{split}
    \textbf{A} = \textbf{A}_{\text{EOM}} &= [\mt{M}_i,\: -\mt{S}^\top], \\
    \textbf{b} =\textbf{b}_{\text{EOM}} &= [-\mt{b}_i -\mt{g}_i + \mt{\tau}_{\text{int}} + \mt{S}^\top\mt{\tau}_\text{friction}],
\end{split}
\end{equation}
where $i \in \{\text{ls}, \text{rs}, \text{ds}\}$. 

Inequality constraints arise from physical and safety constraints on motor torque, power, and velocity. 
Motor torque and power constraints are presented in Equations~\eqref{eq:task_torque} and~\eqref{eq:task_power}, respectively,
\begin{equation}
\begin{split}
    \textbf{D}_{\mt{\tau}_{\text{max}}^+} = [0_{4 \times 5},\: \mathbb{I}_{4\times4}], \; \textbf{f}_{\mt{\tau}_{\text{max}}^+} &= \mt{\tau}_{\text{max}} \\
    \textbf{D}_{\mt{\tau}_{\text{max}}^-} = [0_{4 \times 5},\: -\mathbb{I}_{4\times4}], \; \textbf{f}_{\mt{\tau}_{\text{max}}^-} &= \mt{\tau}_{\text{max}},
    \label{eq:task_torque}
\end{split}
\end{equation}
\begin{equation}
\begin{split}
    \textbf{D}_{\mt{\mathcal{P}}_{\text{max}}^+} = [0_{4 \times 5},\: \mathbb{I}_{4\times4}], \; \textbf{f}_{\mt{\mathcal{P}}_{\text{max}}^+} &= \frac{\mt{\mathcal{P}}_{\text{max}}}{|\dot{\mt{q}}|} \\
    \textbf{D}_{\mt{\mathcal{P}}_{\text{max}}^-} = [0_{4 \times 5},\: -\mathbb{I}_{4\times4}], \; \textbf{f}_{\mt{\mathcal{P}}_{\text{max}}^-} &= \frac{\mt{\mathcal{P}}_{\text{max}}}{|\dot{\mt{q}}|},
    \label{eq:task_power}
\end{split}
\end{equation}
where $\mt{\tau}_{\text{max}} = 80~\text{Nm}$ and $\mt{\mathcal{P}}_{\text{max}} = 100~\text{W}$ are the maximum allowable torque and power of the motors, respectively, \textcolor{black}{and the superscripts $+$ and $-$ are used to indicate the maximum and minimum limits, respectively.}

The joint velocity constraint is converted to an acceleration constraint such that the \textcolor{black}{joint speed} at the next time step will not exceed maximum limits,
\begin{equation}
\begin{split}
    \textbf{D}_{\ddot{\mt{q}}^+} = [0_{4\times1},\: \mathbb{I}_{4\times4},\: 0_{4\times4}], \; \textbf{f}_{\ddot{\mt{q}}^+} &= \frac{\dot{\mt{q}}_{\text{max}} - \dot{\mt{q}}}{\Delta t} \\ \textbf{D}_{\ddot{\mt{q}}^-} = [0_{4\times1},\: -\mathbb{I}_{4\times4},\: 0_{4\times4}], \; \textbf{f}_{\ddot{\mt{q}}^-} &= \frac{\dot{\mt{q}}_{\text{max}} + \dot{\mt{q}}}{\Delta t},
    \label{eq:task_vel}
\end{split}
\end{equation}
where $\dot{\mt{q}}_{\text{max}} = 3.0~\text{rad/s}$ and $\Delta t = 0.003~\text{s}$ is the control loop period. Individual $\textbf{D}$ matrices and $\textbf{f}$ vectors in Equations~\eqref{eq:task_torque}--\eqref{eq:task_vel} are vertically stacked to obtain the overall $\textbf{D}$ matrix and $\textbf{f}$ vector in Equation~\eqref{eq:const_opt}.
The controller is illustrated in Figure~\ref{fig:control_diagram}.

Because we have only a single objective function for this task, a constrained optimization suffices. In the case of multiple objectives with different priorities, a hierarchical optimization framework could be applied~\cite{DarioBellicoso2016, Zimmermann2019}. This allows optimization of lower-priority tasks using freedoms in the design vector that do not affect the objective functions of higher-priority tasks.

\section{Experimental Validation}

Three healthy subjects (one male, two female, 71.6$\pm$12.9~kg, 174$\pm$1.7~cm, 29$\pm$2.6~years) participated in this study to test the proposed interaction torque controller. The link lengths of the exoskeletons were adjusted for each participant so that they were comfortable and their joints were aligned with the exoskeleton's. \textcolor{black}{The adjusted link lengths were also entered into software for automatic modification of the dependent parameters such as $\mt{M}$, $\mt{b}$, and $\mt{g}$.} None of the subjects had used a lower-limb exoskeleton before this study. At the beginning of the experiment, subjects were allowed about two minutes of walking with the exoskeleton on a treadmill for familiarization. \textcolor{black}{Subjects were asked to always hold the hand-rails on the sides of the treadmill for safety and to balance in the frontal plane.} 

We tested the haptic rendering capabilities for zero desired interaction torque under three conditions and nonzero desired interaction torques under two different conditions. In addition to the proposed controller (WECC), the following two conditions were tested:

\paragraph {No-drive} A passive exoskeleton with disassembled drives was used. \textcolor{black}{This allowed us to have a baseline condition where the main sources of friction and inertia (apparent rotor inertia) are eliminated. Subjects still felt the weight of the exoskeleton, however.}
\paragraph {Simplified} The exoskeleton legs are modeled and controlled as independent double pendulums. 
\textcolor{black}{Equations~\eqref{eq:simpleEoM} and~\eqref{eq:simplified_interaction_estimation}  were used to estimate the interaction torques and follow desired accelerations, respectively. The details of this controller are presented in Appendix~\ref{app:simplified}.}

The institutional review board (IRB) of Northwestern University approved this study (STU00212684), and all procedures were in accordance with the Declaration of Helsinki.

\subsection{Evaluation of Transparency}


Haptic transparency was tested for our proposed WECC method, the double-pendulum-based simplified controller, and the passive no-drive condition. For the WECC and simplified controller, zero desired interaction torque was commanded for each generalized coordinate. Each condition was tested for three trials of around \textcolor{black}{65} seconds walking on a treadmill with a speed of 1.1~km/h. \textcolor{black}{The first five seconds of data \textcolor{black}{were} discarded to focus on steady-state walking.}
A relatively slow speed was selected due to the usual walking training speeds in physical rehabilitation settings for \textcolor{black}{individuals with stroke or incomplete spinal cord injury and the joint velocity limits of 3.2~rad/s of the ExoMotus-X2 exoskeleton}.
Trials of the WECC and simplified controllers were performed in a randomized order for each subject. All trials of the passive no-drive condition were performed consecutively for practical reasons, as this condition requires a different exoskeleton with disassembled drives. 
A metronome was played at 40~bpm, and subjects were asked to synchronize their heel strikes with the beats of the metronome to have similar step lengths between subjects and trials.

Human-exoskeleton interaction torques and muscle activity were used to assess transparency performance.

\subsection{Evaluation of Haptic Rendering}


We tested the interaction torque rendering capabilities of the WECC controller and the simplified controller when the desired interaction torques are generated by virtual springs and dampers at the joints,
\begin{equation}
     \mt{\tau}_\text{int}^* = \mt{K}(\mt{q} - \mt{q^*}) + \mt{C}\dot{\mt{q}},
    \label{eq:tau_int_des}
\end{equation}
where $\mt{q^*}$ is the neutral position vector of the virtual springs and $\mt{K} \in \mathbb{R}^{5\times 5}$ and $\mt{C} \in \mathbb{R}^{5\times 5}$ are diagonal stiffness and damping matrices, respectively. A constant $q^*_i$ was used at each joint $i$, and the desired interaction torque at the backpack was set to zero by assigning $K_0$ and $C_0$ to zero. \textcolor{black}{The constant neutral angles were chosen such that the desired interaction torques have a similar magnitude in both directions.} Since the motion of a leg during stance is slower and has a smaller range, larger stiffness and damping constants were used for stance phases than for swing phases. \textcolor{black}{The magnitudes of the stiffness and damping were chosen during pilot trials such that the rendered torque is significant but does not prevent the user from walking at the requested speed.} The diagonal values of the stiffness ($k_i$) and damping ($c_i$) matrices, as well as $\mt{q^*}$, are presented in Table~\ref{tab:rendering_params}. To avoid discontinuities in the desired interaction torque during stance to swing transitions, the desired interaction torque was low-pass filtered with a cutoff frequency of 5~Hz.
\begin{table}
\begin{center}
\caption{Virtual stiffness and damping values used for haptic rendering evaluation.}\vspace{1ex}
    \label{tab:rendering_params}
\begin{tabular}{cc|ccc}
\cline{1-5}
\rule{0pt}{2ex}
Joint & Leg Phase & $k_i$ [Nm/rad] & $c_i$ [Nms/rad] & ${q_i}^*$ [degrees]     \\ 
\cline{1-5}
Hip         & Stance & 50 & 10 & 25      \\
Hip         & Swing & 30 & 6 & 25      \\
Knee         & Stance & 30 & 6 & -45     \\
Knee        & Swing & 25 & 5 & -45     \\ \cline{1-5}
\end{tabular}
\end{center}
\end{table}
\subsection{Data Recording and Analysis}

Joint positions, velocities, and interaction torques were recorded during all trials. The exoskeleton without drives is instrumented with IMUs (Tech-IMU V4, Technaid, Spain) at each link to measure joint motions, which are used in interaction torque calculation. \textcolor{black}{Because IMUs were placed on the exoskeleton and not on the user, they were not affected by possible human-robot joint misalignment.} All exoskeleton data were collected at 333~Hz.

Nine bipolar electromyography (EMG) electrodes (MA400 EMG Systems, Motion Lab Systems, USA) were placed on the belly of the following muscles for the evaluation of transparency experiments: rectus femoris (RF), vastus lateralis (VL), vastus medialis (VM), biceps femoris (BF), semitendinosus (ST), medial gastrocnemius (MG), lateral gastrocnemius (LG), and soleus (SOL). The electrodes were placed only on the left leg \textcolor{black}{and not removed between trials}. EMG data were recorded at 1500~Hz. The raw EMG signals were first bandpass filtered between 20 and 500~Hz with a sixth-order Butterworth filter. Then, a notch filter between 59 and 61~Hz was applied to reduce noise due to power line interference. Finally, the signals were low-pass filtered at 10~Hz after full-wave rectification. Processed EMG signals were normalized with respect to the mean of the no-drive condition for each subject independently. For one subject, RF recordings were not successfully collected due to sensor failure.

Exoskeleton-based measurements and processed EMG readings were windowed between the heel strikes of the corresponding leg. Windowed data from the same joint on both legs and three subjects were lumped together for each condition. The mean absolute value of the whole cycle and only stance or swing phases for each gait cycle were used as sample points and visualized in the corresponding box plots in the Results section. 

\textcolor{black}{To test the difference in means between each condition for each measure (i.e., interaction torque, muscle activity), we used two-way ANOVA analyses with condition and subject as fixed effects, as well as the interaction between these variables. Post-hoc comparisons were made using Tukey’s Honestly Significant Difference (HSD) test for transparency experiments comparing the three conditions. A $p$-value of 0.05 was used to accept or reject null hypotheses. Though the effects of inter-subject variability were included in the ANOVA analyses, we only present the ANOVA results and post-hoc comparisons for the condition, as the focus of this paper is controller development.} To ensure consistency with ANOVA analyses and prevent the mixing of inter- and intra-subject variability, we removed the inter-subject variability from the data presented in the box plots and Tables in the following section. This is achieved by demeaning the data of each condition/subject pair and subsequently adding the overall mean of each condition, including three subjects.


\section{Results}
\subsection{Transparency}

\begin{figure*}
\centering
    \includegraphics[height=0.65\textwidth]{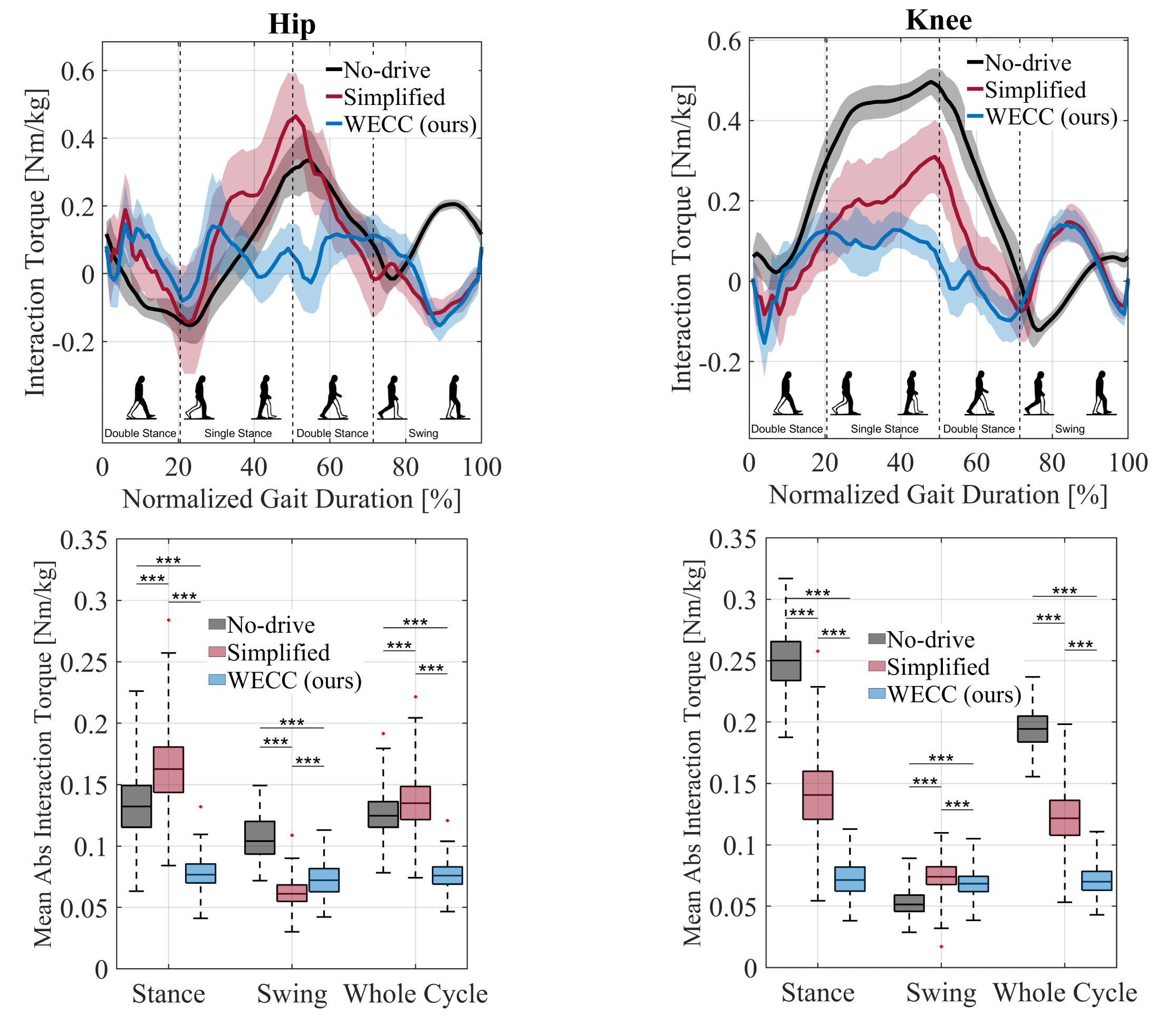}
\caption{Interaction torques between the human and exoskeleton during transparency trials. The figures on the top row show the interaction torques vs. normalized gait duration \textcolor{black}{for a representative subject}. The data includes every step of both legs, and the shaded area represents $\pm$ standard deviation. On the bottom row, the box and whisker plots of the mean absolute interaction torques during the stance phase, swing phase, and the whole gait cycle are presented. Mean absolute values of the corresponding phase at each step of both legs of all three subjects are used as a data point \textcolor{black}{after removing the inter-subject variability} in box and whisker plots ($N \approx 320$). \textcolor{black}{\textsf{***} indicates $p < 0.001$ from Tukey's HSD test.}} 

\label{fig:trans_torque}
\end{figure*}

\textcolor{black}{Participants were able to follow the metronome, and there was no significant difference between the step time at different conditions (No-drive: 1.49 $\pm$ 0.06~s, Simplified: 1.51 $\pm$ 0.06~s, WECC: 1.50 $\pm$ 0.06~s).} 
Interaction torques during the transparency trials, normalized by the body mass of each subject (units of Nm/kg), are presented in Figure~\ref{fig:trans_torque} and Table~\ref{tab:trans_torque_result}.
Below we describe the interaction torques during a leg's swing phase, its stance phase (both single stance and double stance unless otherwise noted), and its entire cycle.


The performance of the proposed WECC controller was consistent throughout the whole gait cycle with a mean absolute interaction torque around 0.07 $\pm$ \textcolor{black}{0.02}~Nm/kg both at the hip and knee joints during both stance and swing phases. The simplified controller resulted in a similar mean absolute interaction torque during the swing phase (difference of mean absolute interaction torques~$\Delta\leq$0.01~Nm/kg) but significantly higher mean absolute interaction torque of 0.16 and 0.14~Nm/kg at the hip and knee joints, respectively, during the stance phase. During the stance phase, higher interaction torques were observed for the no-drive condition than for WECC at both the hip ($\Delta \geq$ 0.06~Nm/kg, $p < 0.001$) and knee ($\Delta \geq$ 0.17~Nm/kg, $p < 0.001$) joints. During the swing phase, the no-drive condition resulted in  significantly higher interaction torques  at the hip joint ($\Delta \geq$ 0.05~Nm/kg, $p < 0.001$) and similar interaction torques at the knee joint ($\Delta \approx$ 0.01~Nm/kg) compared to the proposed WECC method. Average transparency performance during the whole cycle was best for the WECC controller, with mean absolute interaction torques of 0.076 $\pm$ \textcolor{black}{0.011}~Nm/kg and 0.071 $\pm$ \textcolor{black}{0.011}~Nm/kg at the hip and knee joints, respectively. The simplified controller achieved better mean absolute interaction torque than the no-drive condition at the knee joint ($\Delta \approx$ 0.07~Nm/kg) averaged over the whole gait cycle. At the hip joint, both performances were similar with a mean interaction torque difference around 0.01~Nm/kg.

\begin{table}
\begin{center}
\caption{Mean and (standard deviation) of the interaction torques between the human and exoskeleton at the hip and knee joints during transparency experiments. All values are expressed in Nm/kg. \textcolor{black}{$p$-values for the effect of condition in the two-way ANOVA and significant post-hoc comparisons are presented in the right columns.} }\vspace{1ex}
    \label{tab:trans_torque_result}
\setlength{\tabcolsep}{3pt} 
\begin{tabular}{c|ccc|cc}
\hline
Joint/Phase & \makecell{No-drive \\ ($c_1$)} & \makecell{Simplified \\ ($c_2$)} & \makecell{WECC \\ ($c_3$)} & \textcolor{black}{$p$} & \textcolor{black}{Tukey's HSD}   \\ 
\hline
Hip/Stance & 0.13 (0.03) & 0.16 (0.03) & 0.08 (0.01) & $<$0.001 & $c_3<c_1<c_2$     \\
Hip/Swing & 0.11 (0.02) & 0.06 (0.01) & 0.07 (0.01) & $<$0.001 & $c_2<c_3<c_1$    \\
Hip/WC & 0.13 (0.02) & 0.14 (0.02)& 0.08 (0.01) & $<$0.001 & $c_3<c_1<c_2$    \\
Knee/Stance & 0.25 (0.02) & 0.14 (0.03) & 0.07 (0.01) & $<$0.001 & $c_3<c_2<c_1$  \\
Knee/Swing & 0.001 (0.01) & 0.07 (0.01) & 0.07 (0.01) & $<$0.001 & $c_1<c_3<c_2$    \\
Knee/WC & 0.19 (0.02) & 0.12 (0.02) & 0.07 (0.01) & $<$0.001 & $c_3<c_2<c_1$    \\
\hline
\end{tabular}
\end{center}
\end{table}

\begin{table}
\begin{center}
\caption{Mean and (standard deviation) of muscle activities during transparency experiments, normalized with respect to the mean of the no-drive condition. \textcolor{black}{$p$-values for the effect of condition in the two-way ANOVA and significant post-hoc comparisons are presented in the right columns.}}\vspace{1ex} 
    \label{tab:trans_emg_result}
\setlength{\tabcolsep}{3pt} 
\begin{tabular}{c|ccc|cc}
\hline
Muscle & \makecell{No-drive \\ ($c_1$)} & \makecell{Simplified \\ ($c_2$)} & \makecell{WECC \\ ($c_3$)} & \textcolor{black}{$p$} & \textcolor{black}{Tukey's HSD}     \\ 
\hline
RF & 1 (0.19) & 0.83 (0.14) & 0.77 (0.14) & $<$0.001 & $c_3<c_2<c_1$     \\
VL & 1 (0.19) & 0.64 (0.15) & 0.59 (0.09)  & $<$0.001 & $c_3<c_2<c_1$     \\
VM & 1 (0.21) & 0.59 (0.14) & 0.56 (0.09) & $<$0.001 & $c_1>c_2, c_3$     \\
BF & 1 (0.19) & 1.32 (0.29) & 1.94 (0.36)  & $<$0.001 & $c_1<c_2<c_3$     \\
ST & 1 (0.28) & 1.57 (0.37) & 2.51 (0.41)  & $<$0.001 & $c_1<c_2<c_3$      \\
TA & 1 (0.23) & 0.97 (0.20) & 1.06 (0.15)  & $<$0.001 & $c_3>c_1, c_2$      \\
LG & 1 (0.18) & 1.56 (0.31) & 1.91 (0.33) & $<$0.001& $c_1<c_2<c_3$    \\
MG & 1 (0.20) & 1.58 (0.28) & 1.95 (0.33) & $<$0.001 & $c_1<c_2<c_3$      \\
SOL & 1 (0.17) & 0.76 (0.15) & 0.72 (0.09) & $<$0.001 & $c_3<c_2<c_1$     \\
\cline{1-6}
\end{tabular}
\end{center}
\end{table}

\begin{figure*}
\centering
    \includegraphics[height=0.45\textwidth]{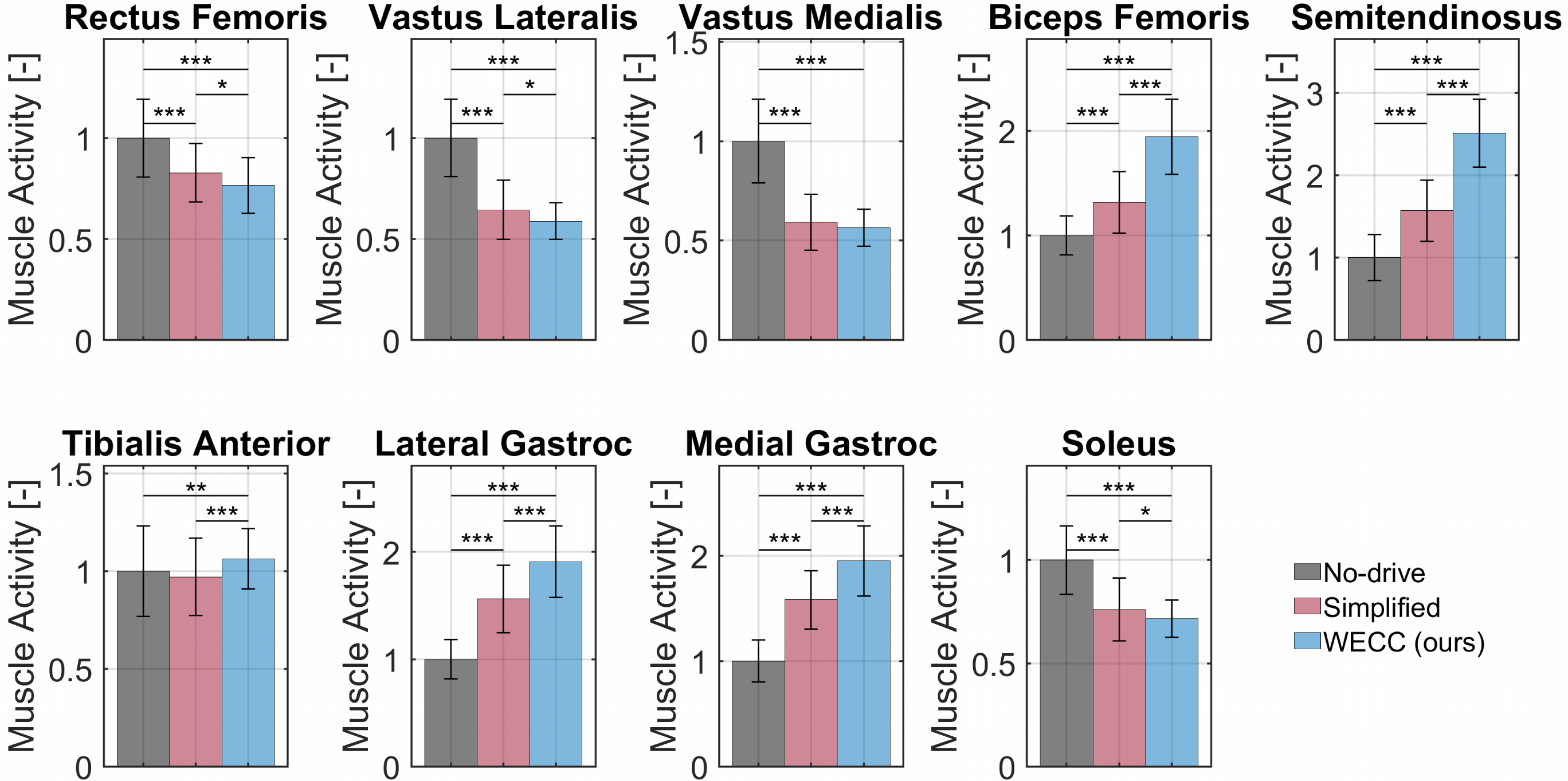}
\caption{Muscle activity during transparency experiments. Data are normalized with respect to the mean of the no-drive condition. Mean absolute values of the filtered EMG data averaged over the whole cycle at each step of all subjects are used \textcolor{black}{after removing the inter-subject variability} as a data point in bar plots ($N \approx 160$). \textcolor{black}{\textsf{*}, \textsf{**}, and \textsf{***}  indicate $p < 0.05$, $0.01$, and $0.001$, respectively, from Tukey's HSD test.}}
\label{fig:trans_emg}
\end{figure*}

Figure~\ref{fig:trans_emg} and Table \ref{tab:trans_emg_result} show the muscle activity of the various muscles during the transparency trials. It is seen that muscles responsible for hip flexion (RF) and knee extension (RF, VL, VM) had the lowest activation with the WECC controller and highest with the passive no-drive condition. On the other hand, our controller resulted in the most activity, and the no-drive condition resulted in the least activity, for the muscles responsible for hip extension and knee flexion (BF, ST). Furthermore, the SOL muscle, responsible for ankle plantar flexion, was used the least with the WECC controller and the most under the no-drive condition. In contrast, LG and MG muscles, responsible for knee flexion and ankle plantar flexion, had the highest activation with the WECC controller and the lowest under the no-drive condition. 

\begin{table}
\begin{center}
\caption{Mean and (standard deviation) of the interaction torque error between the human and exoskeleton at hip and knee joints during haptic rendering experiments. All values are expressed in Nm/kg. \textcolor{black}{$p$-values for the effect of condition in the two-way ANOVA are presented in the right column.}}\vspace{1ex}
    \label{tab:rendering_torque_result}
\setlength{\tabcolsep}{3pt} 
\begin{tabular}{c|cc|c}
\hline
Joint/Phase & \makecell{Simplified \\ ($c_2$)} & \makecell{WECC \\ ($c_3$)} & \textcolor{black}{$p$}    \\ 
\hline
Hip/Stance & 0.12 (0.02) & 0.05 (0.01)  & $<$0.001      \\
Hip/Swing & 0.04 (0.01) & 0.04 (0.01) & 0.49     \\
Hip/WC& 0.10 (0.02) & 0.05 (0.01) & $<$0.001     \\
Knee/Stance & 0.21 (0.02) & 0.05 (0.01) & $<$0.001   \\
Knee/Swing & 0.04 (0.01) & 0.05(0.01) & $<$0.001    \\
Knee/WC & 0.17 (0.02) & 0.05 (0.01) & $<$0.001    \\ \cline{1-4}
\end{tabular}
\end{center}
\end{table}

\subsection{Haptic Rendering}

\begin{figure}
\hspace*{-0.4cm} \centering
    \includegraphics[height=0.25\textwidth]{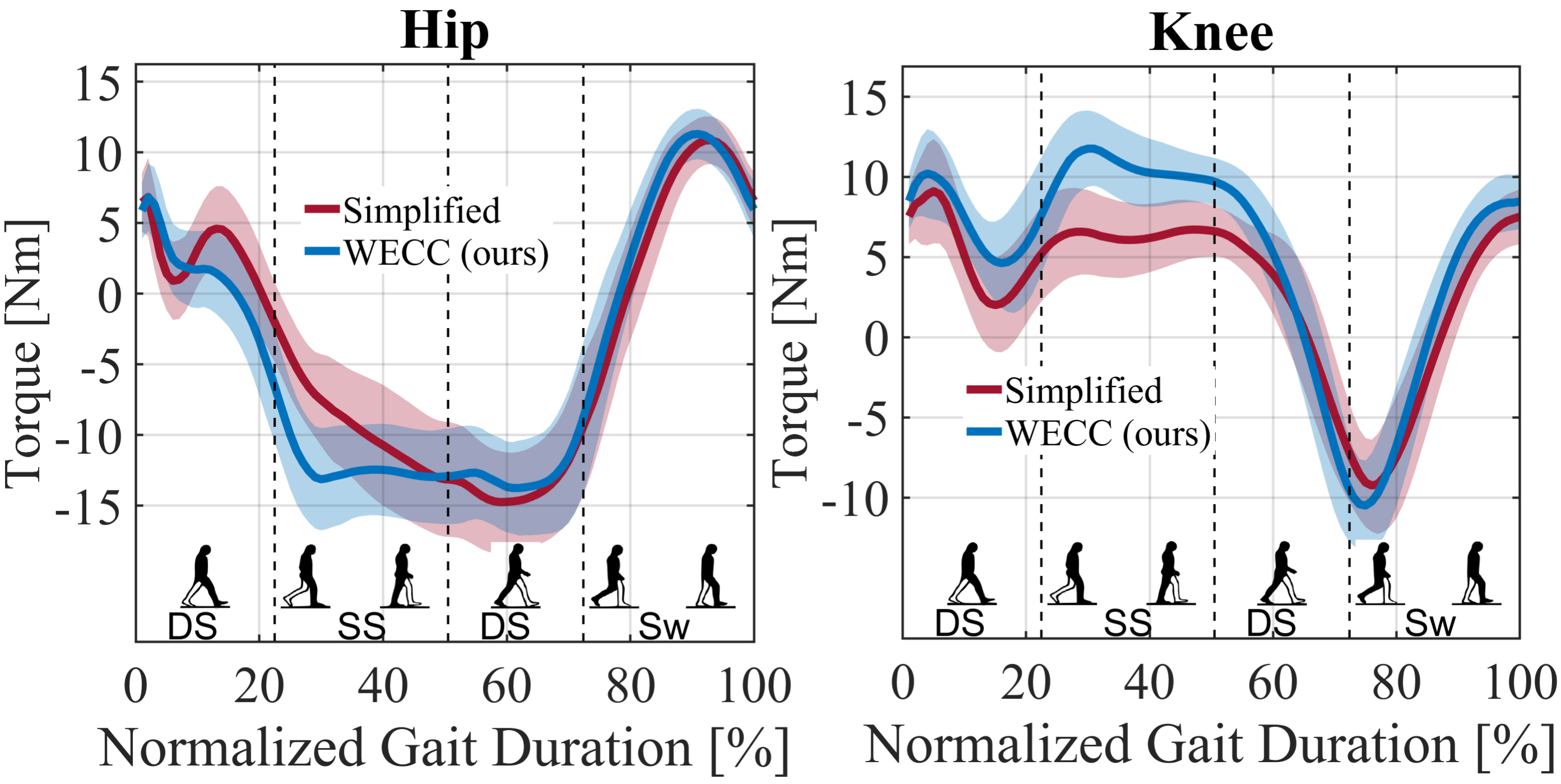}
\caption{Desired interaction torque due to the rendered spring damper elements.}
\label{fig:render_desired_torques}
\end{figure}

Properties of the rendered virtual spring-damper elements shown in Table~\ref{tab:rendering_params} resulted in the desired interaction torques presented in Figure~\ref{fig:render_desired_torques}. The figure includes data from the joints of both legs and three subjects. Desired interaction torques vary in the range $\pm$15~Nm during the gait cycle for all subjects. \textcolor{black}{There was no significant difference between the step times at different conditions (Simplified: 1.32 $\pm$ 0.13~s, WECC: 1.31 $\pm$ 0.13~s). The demonstration of the haptic rendering performance of the WECC control is also presented in this video \footnote{\url{https://tinyurl.com/ExoHapticRendering}}.}

\begin{figure*}
\centering
    \includegraphics[height=0.65\textwidth]{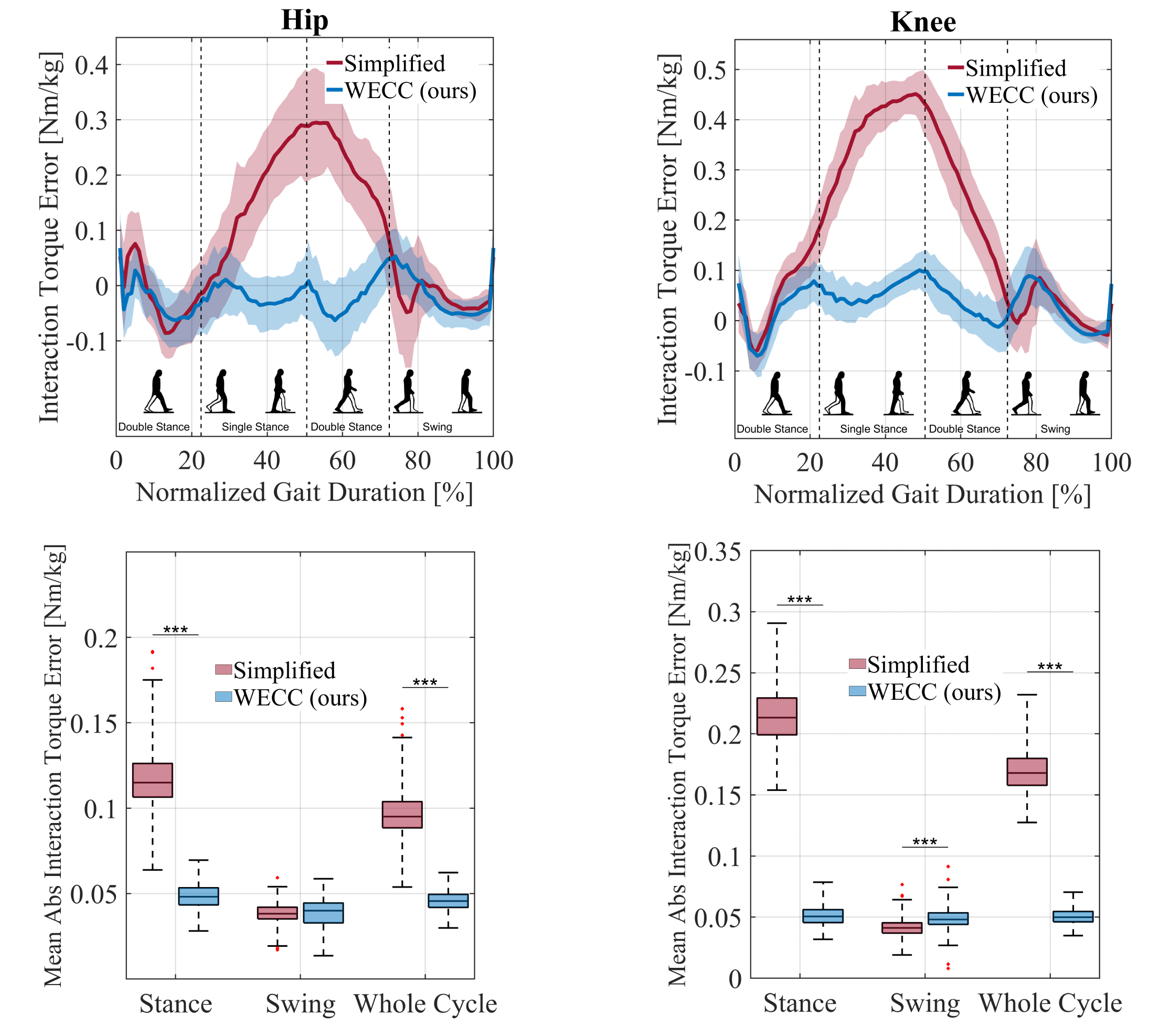}
\caption{Interaction torque error during haptic rendering trials. The figures on the top row show the interaction torque error vs. normalized gait duration \textcolor{black}{for a representative subject}. The data includes every step of both legs, and the shaded area represents $\pm$ standard deviation. On the bottom row, the box and whisker plots of the mean absolute interaction torque error during the stance phase, swing phase, and the whole gait cycle are presented. Mean absolute values of the corresponding phase at each step of both legs of all three subjects are used \textcolor{black}{after removing the inter-subject variability} as a data point in box and whisker plots ($N \approx 370$). \textcolor{black}{\textsf{***} indicates $p < 0.001$ from ANOVA results.}}
\label{fig:render_torques}
\end{figure*}

The rendering performance of our proposed WECC controller and the simplified controller are shown in Figure~\ref{fig:render_torques} and Table~\ref{tab:rendering_torque_result}.
Similar to the performance of transparency trials, our controller's interaction torque tracking error was consistent throughout the whole cycle.
On the other hand, the simplified controller resulted in a large peak of interaction torque errors on both joints
near the end of the single stance state.

The WECC controller resulted in a normalized mean absolute interaction torque error of 0.049 $\pm$ \textcolor{black}{0.008}~Nm/kg and 0.039 $\pm$ \textcolor{black}{0.008}~Nm/kg at the hip joint during the stance and swing phases, respectively. The simplified controller resulted in a similar mean absolute error during the swing phase both at the hip and knee joints ($\Delta \leq 0.01$~Nm/kg). On the other hand, the average absolute interaction torque error was significantly larger during the stance phase both at the hip ($\Delta \geq 0.07$~Nm/kg, $p<0.05$) and knee ($\Delta \geq 0.16$~Nm/kg, $p<0.05$) joints. Averaged over the whole cycle, the WECC controller resulted in an error of 0.050 $\pm$ \textcolor{black}{0.006}~Nm/kg for both hip and knee joints. The errors of the simplified controller were significantly larger both at the hip ($\Delta \geq 0.05$~Nm/kg, $p<0.05$) and knee joints ($\Delta \geq 0.12$~Nm/kg, $p<0.05$).


\section{Discussion}
\label{sec:Discussion}
\subsection{Interaction Torque Tracking}

The simplified controller and the proposed WECC controller performed similarly during the swing phase for both joints in the transparency and haptic rendering trials, as shown in Figures~\ref{fig:trans_torque} and~\ref{fig:render_torques}. Since the swing leg is not affected by the weight of the backpack and the other leg, the two-link simplified model is sufficient and results in accurate interaction torque estimation and motor torque calculation for the swing leg. On the other hand, the simplified controller failed to render the desired interaction torques for the stance leg. This is because the model does not consider whole-body dynamics, resulting in incorrect interaction torque estimation for the stance leg. For the simplified controller, subjects feel additional torques on their joints due to the uncompensated weight of the exoskeleton. \textcolor{black}{The calculation of dynamical parameters and the differences between the two approaches are presented in Appendi\textcolor{black}{ces}~\ref{app:dyn_params} and~\ref{app:differences}.}
As the proposed WECC controller uses whole-body dynamics and considers physical limitations, the interaction torque tracking performance was consistent throughout the entire gait cycle. 

The most transparent performance at the knee joint was obtained during the swing phase of the no-drive condition. This is because the weight of the shank is insignificant, and there is low friction and no apparent rotor inertia. 
Because of the larger inertia, the no-drive condition results in more interaction torque at the hip joint compared to the active controllers, where the weight of the leg is compensated. The highest amount of interaction torque was observed for the stance leg under the no-drive condition, as the stance leg of the subject needs to overcome the torques due to the weight of the exoskeleton's stance leg, swing leg, and backpack.

\textcolor{black}{The proposed method uses the exoskeleton model to estimate interaction torque and control acceleration to track the desired interaction torque. Considering the model-based dynamics based on the gait state results in consistent performance throughout the whole gait cycle. This model-based approach depends on an accurate dynamic model. Thanks to CAD tools, this is not difficult to obtain. Accurate friction modeling is more challenging. Performance can be improved by (1) using a more accurate friction model and (2) estimating joint accelerations in real time and implementing an inner acceleration control loop to compensate modeling errors. Estimating real-time accelerations also allows considering the neglected inertial terms in interaction torque estimation. In our post-experiment analyses, we observed that the mean absolute values of the neglected inertial torques were around 14\% of the total dynamical forces.}

\subsection{Muscle Activity}

The muscles responsible for hip flexion (RF) and knee extension (RF, VL, VM) were activated the least with the proposed WECC controller and the most with the passive no-drive condition during the transparency trials. This is mainly due to the accurate compensation of the exoskeleton dynamics for the WECC controller.


\textcolor{black}{
The opposite trend was observed for the muscles responsible for hip extension and knee flexion (BF, ST). These muscles were activated more with the WECC controller than for the no-drive and simplified conditions. This is expected for BF and ST, remembering that the transparency goal for WECC is zero interaction torque, not assist. The uncompensated weight of the exoskeleton in the no-drive and simplified conditions results in significant hip-extension torque assist during early double stance, as shown in Figure 9 by the large positive power flow from the exoskeleton to the user.
}

\begin{figure}
\centering
    \includegraphics[height=0.35\textwidth]{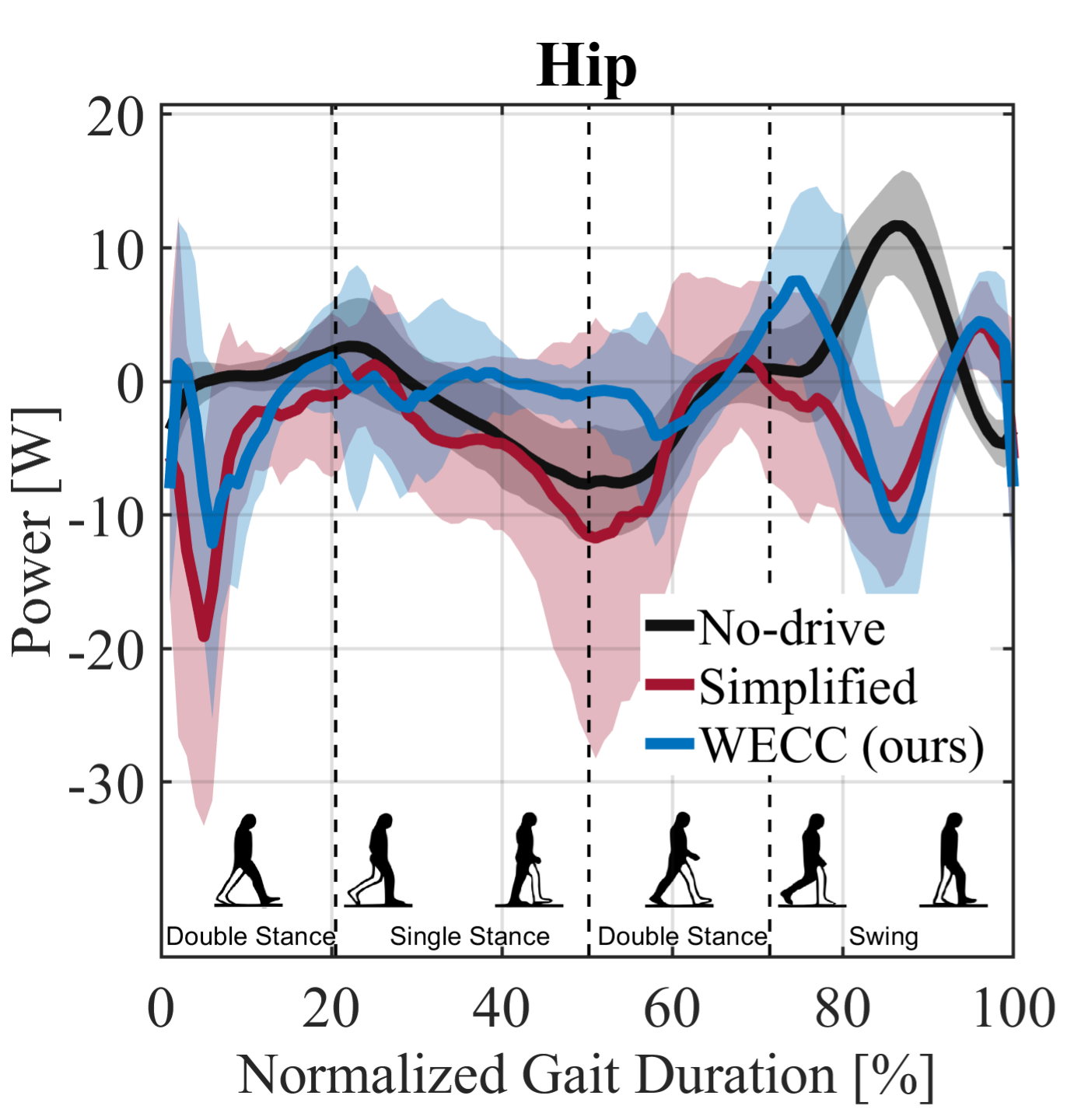}
\caption{Hip interaction power during transparency experiments. Positive and negative values indicate resistance and assistance to the human, respectively.}
\label{fig:power}
\end{figure}

The gastrocnemius muscles (LG, MG), which cause ankle plantar flexion and knee flexion, were activated most with the WECC controller and least with the no-drive condition. On the other hand, an opposite trend was observed for the SOL muscle, which causes only ankle plantar flexion. High interaction torques in the knee flexion direction (positive direction in Figure~\ref{fig:trans_torque}) for the simplified and no-drive conditions leads to less gastrocnemius activity in these conditions. This explains the different trend between gastrocnemius and SOL muscles.

\subsection{\textcolor{black}{Overground Walking}}
\label{ssec:overground_walking}
\textcolor{black}{
The proposed method requires the ratios of the vertical GRF and the direction of GRF in the sagittal plane. While we used external force plates under the treadmill, these terms can also be obtained using wearable sensors or sensors instrumented onto the exoskeleton. For example, the ratio of the vertical forces can be measured with simple and cost-effective FSR sensors or pressure pads.}

\begin{figure}
\centering
\hspace*{-0.4cm}    \includegraphics[height=0.24\textwidth]{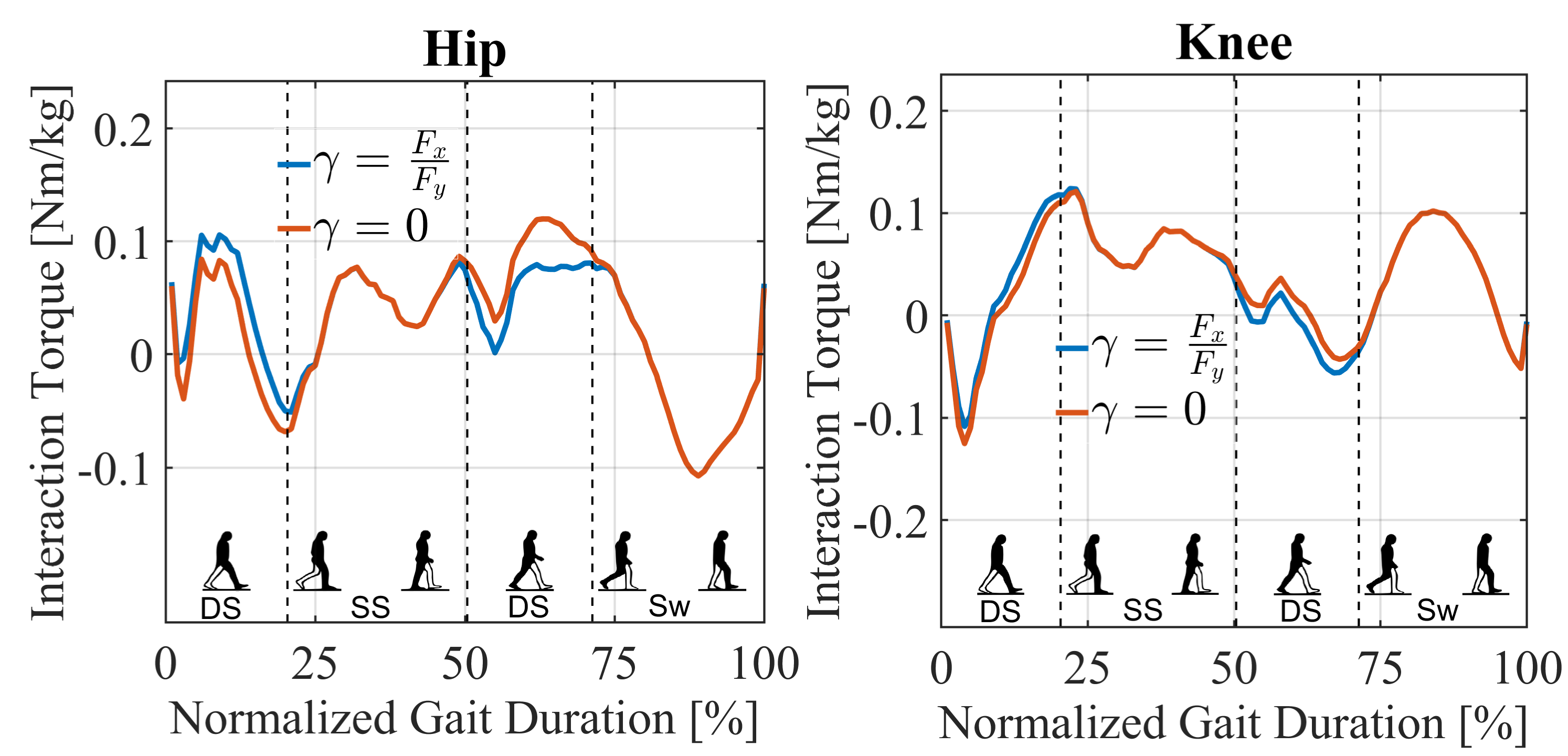}
\caption{Mean interaction torque estimation of all subjects during the all WECC trials using the direction of the GRF ($\gamma = F_x/F_y$) and assuming horizontal GRF is negligible ($\gamma = 0$).}
\label{fig:gamma0_comparison}
\end{figure}

\textcolor{black}{
It is relatively harder to measure the horizontal GRF because instrumented insoles usually provide only vertical forces. One can implement force sensors that provide two-directional force measurements~\cite{Jang2022} or indirectly estimate the anterior-posterior force assuming the foot does not slip on the ground~\cite{Gehlhar2022} or using a model and estimated accelerations~\cite{Revi2020}. 
Because accelerations and decelerations are small for slow gait speeds, it is also possible to assume horizontal GRF to be near zero. The effect of this assumption on the interaction torque estimation was tested on the collected experimental data from all subjects during WECC conditions. Figure \ref{fig:gamma0_comparison} shows that the interaction torque estimates are changed little by this assumption.}

\begin{figure}
\centering
\hspace*{-0.4cm}    \includegraphics[height=0.24\textwidth]{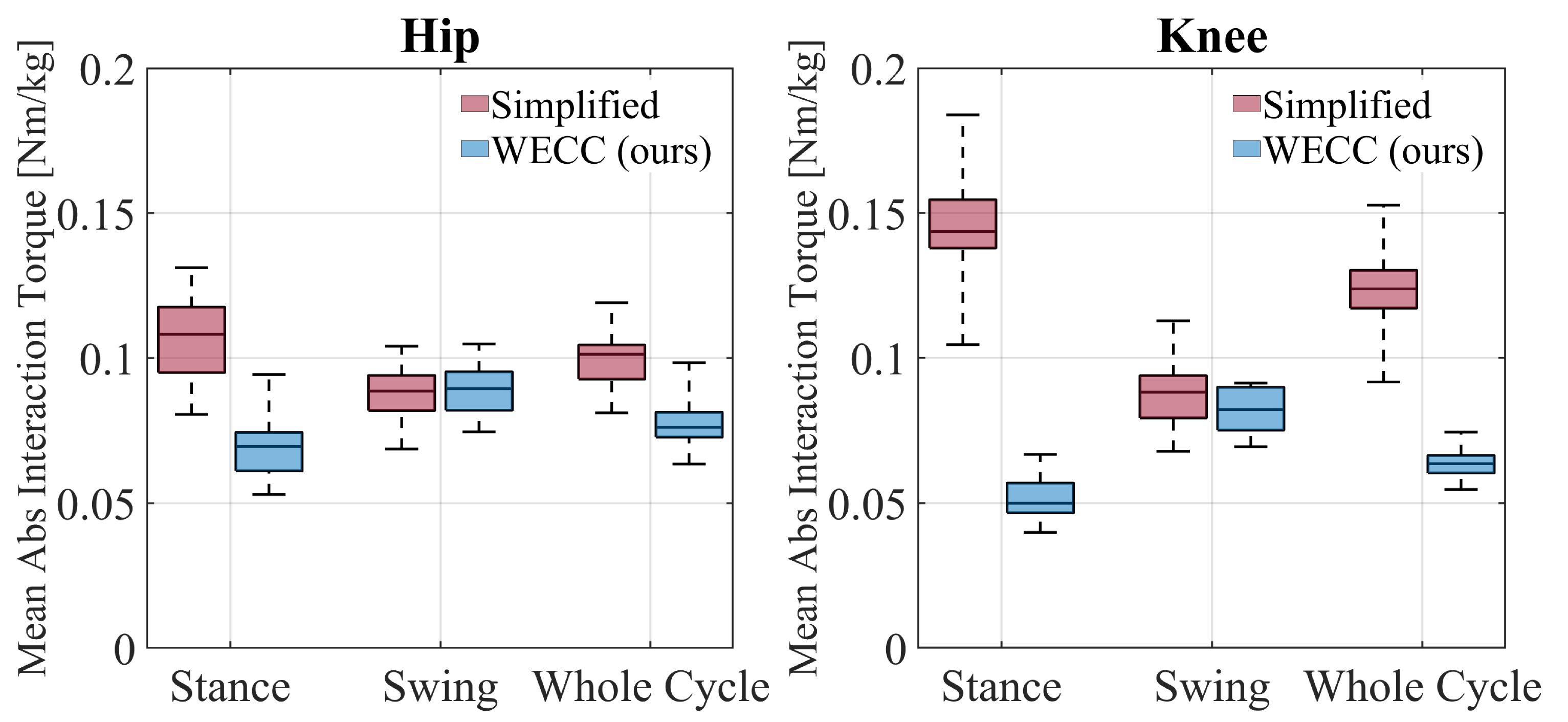}
\caption{Mean absolute interaction torques during the pilot overground tests.}
\label{fig:overground_results}
\end{figure}

\textcolor{black}{
The supplementary video includes preliminary overground tests with FSR pads implemented at the bottom of the feet while assuming horizontal forces to be near zero. The user in the video climbs over a small step and walks forward and backward. \textcolor{black}{We also conducted a pilot test where a user walked a distance of 4.5 m four times with each controller (WECC and simplified) in haptic transparent mode, while holding onto parallel bars.
We observed similar trends between this overground walking test and the treadmill experiments. \textcolor{black}{Mean absolute interaction torque results are presented in Figure~\ref{fig:overground_results}.} In the stance phase, the WECC controller resulted in a 35\% and 65\% decrease in mean absolute interaction torque at the hip and knee joints, respectively, compared to the simplified controller. The performance of the two controllers was similar in the swing phase (difference less than 1\% and 5.5\% for hip and knee, respectively). }
}

\textcolor{black}{
}

\section{Conclusion}

We have presented the whole-exoskeleton closed-loop compensation (WECC) controller to measure and control the interaction torques between a user and a floating base hip-knee exoskeleton that contacts the ground. The WECC controller calculates the interaction torques using joint torque measurements and gait state information. The desired joint interaction torques are achieved by using a virtual mass controller together with a constrained optimization scheme that considers the whole-exoskeleton dynamics, physical limits, and safety constraints. The performance of the WECC controller was compared in terms of transparency and spring-damper haptic rendering with a commonly-used simplified double-pendulum model where the legs are modeled independent of one another. WECC control resulted in consistent interaction torque tracking during the whole gait cycle for both zero and nonzero desired interaction torques, while the simplified controller failed to track desired interaction torques during the stance phase.

Future improvements to WECC control could include explicit modeling of joint limits in the dynamic model, particularly because the knee joint gets close to the physical limit during the gait cycle. This will improve the accuracy of the forward model and therefore the tracking of the optimized accelerations. Moreover, an inner acceleration loop to compensate for the modeling inaccuracies, using additional IMUs or reinforcement learning -based strategies~\cite{WenOnline2020} for online optimization of system parameters, may be options to improve the current results. 

\textcolor{black}{A limitation when comparing interaction torques between different controllers arises from the challenge of obtaining ground-truth data. Reflected interaction torques are influenced by forces at each contact point, necessitating the installation of F/T sensors at every contact point for the measurement of these interaction torques. Because this is practically challenging, we used the whole-body dynamics of the exoskeleton to estimate the interaction torques as accurately as possible while comparing different controller performances. More accurate comparisons could be possible with an experimental setup where the contact points are kept at a minimum and all are sensorized.}

The WECC controller can \textcolor{black}{potentially be used to adjust the interaction torques to assist a user in overground walking, to resist for training in a clinical setting, or to provide haptic transparency on lower-limb exoskeletons with feet. Commercial versions of this type of exoskeleton are usually heavy, have high friction, and do not have force-torque sensors. Consequently, they have been utilized primarily for position control rather than accurate interaction force control. The proposed WECC control overcomes these limitations and} is especially beneficial for heavy lower-limb exoskeletons where the whole-body dynamics need to be compensated. In future studies, this infrastructure will be used to investigate the effects of dyadic haptic interaction on performance and learning for complex multi-DOF lower-limb motions~\cite{Kucuktabak2021, Kim2020, Kim2022}. 

\appendix

\color{black}
\subsection{Simplified Controller}
\label{app:simplified}

\begin{figure*}
\centering
\begin{tikzpicture}[->,>=stealth',shorten >=1pt,auto,node distance=3cm,
                    semithick]
                    
  \tikzstyle{block}=[rectangle, draw=black, text=black, minimum size=1cm, align=center]
  \tikzstyle{node}=[circle, draw=black, text=black, align=center]
  \tikzstyle{empty}=[text=black, align=center]

  \node[block] (MV)    {$\mt{M}_\text{virt}^{-1}$};
  \node[node] (refCircle)  [left = 1 cm of MV]  {-};
  \node[block] (HO) [right = 1.5 cm of MV] {
    \includegraphics[width=.2\textwidth]{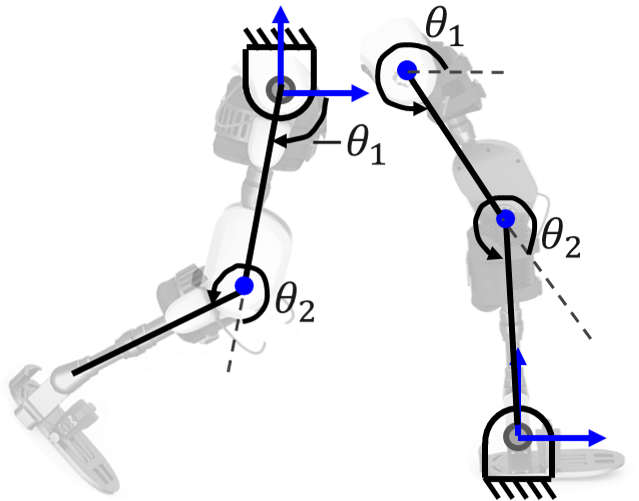}
  };
  \node[empty] (HO_empty) [above = -0.05 cm of HO] {2-link feed-forward model (Eq. \ref{eq:simpleEoM})};
  \node[block] (HR) [right = 2 cm of HO] {
  \includegraphics[width=.1\textwidth]{figures/X2_cartoon.png}
  };
 \node[empty] (HR_empty) [above = -0.05 cm of HR] {Coupled Human-Robot};
  \node[empty] (HR_right) [right = 0.5 cm of HR] {};
  \node[block] (GS) [below = 0.75 cm of HO] {
  \begin{tikzpicture}[->,>=stealth',shorten >=1pt,auto,node distance=3cm,
                    semithick]
  \tikzstyle{every state}=[draw=black, text=black, minimum size=1cm, align=center]

  \node[state] (ls)                    {LS\\ $\alpha = 1$};
  \node[state] (ds) [below right = 0.5 cm of ls] {DS\\ $\alpha \in [0 \ 1]$};
  \node[state] (rs) [above right = 0.5 cm of ds] {RS\\ $\alpha = 0$};

  \path (ls) 
             edge                 node[sloped] {} (ds)
        (rs) 
             edge   [bend left] node[sloped, below] {} (ds)
        (ds) 
            edge  [bend left]  node[sloped, below] {} (ls)
            edge   node[sloped] {} (rs);
\end{tikzpicture}
  };
  \node[empty] (GS_right) [right = 4.3 cm of GS] {};
  \node[block] (FE) [below = 0.5 cm of GS] {Simplified Interaction Torque Estimator (Eq. \ref{eq:simplified_interaction_estimation})};
  \node[block] (FE_wb) [below = 0.9 cm of FE] {Whole-Body Interaction Torque Estimator};
  \node[empty] (FE_wb_left) [left = 1 cm of FE_wb] {};
  \node[empty] (FE_wb_top_left) [above = 0.5 cm of FE_wb_left] {};
  \node[empty] (FE_right) [right = 3.18 cm of FE] {};
  \node[empty] (FE_left) [left = 3.53 cm of FE] {};
  \node[empty] (FE_wb_right) [right = 3.65 cm of FE_wb] {};
  \node[empty] (tau_des) [left = 1 cm of refCircle] {};
  \node[block] (plot) [left = 1.7 cm of FE_wb] {
  \includegraphics[width=0.05\textwidth]{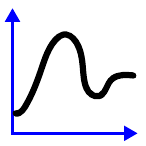}
  };
   \node[empty] (plot_empty) [above = -0.05 cm of plot] {Visualization};
  
  \path (MV) edge node[] {$\ddot{\mt{q}}^*$} (HO);
  \path (HO) edge node[] {$(\ddot{\mt{q}}, \mt{\tau}_{\text{motor}})$} (HR);
  \path (GS) edge node[] {$\alpha$} (HO);
  \path (GS) edge node[] {$\alpha$} (FE);
  
  \draw [-] (HR.east) -| node[pos=0.5] {} (GS_right.center);
  \draw [->] (GS_right.center) -- node[above, pos=0.5] {$\mt{F}^\text{grf}$} (GS);
  \draw [->] (GS.west) -| node[near end] {$\alpha$} (MV.south);
   \draw [-] (GS_right.center) -- node[pos=0.5] {} (FE_right.center);
   \draw [->] (FE_right.center) -- node[above, pos=0.5] {$[\mt{q},\dot{\mt{q}}, \mt{\tau}_\text{joint}]$} (FE);
   \draw [-] (FE_right.center) -- node[above, pos=0.5] {} (FE_wb_right.center);
   \draw [->] (FE_wb_right.center) -- node[above, pos=0.5] {$[\mt{q},\dot{\mt{q}}, \mt{\tau}_\text{joint}, \mt{F}^\text{grf}]$} (FE_wb.east);
   \draw [->, name path=tau_int] (FE.west) -| node[pos = 0.9] {$\mt{\tau}_\text{int}$} (refCircle.south);
   \draw [->] (refCircle) -- node[pos=0.5] {$\mt{\tau}_\text{int}^{\text{err}}$} (MV);
   \draw [->] (tau_des) -- node[pos=0.5] {$\mt{\tau}_\text{int}^{\text{*}}$} (refCircle);
   \draw [->] (FE_wb.west) -> node[pos=0.5] {} (plot.east);

    \path [name path=GStoFE_wb_top_left] (GS.200) -| (FE_wb_top_left);
      
    \path [name intersections={of = tau_int and GStoFE_wb_top_left}];
    \coordinate (S)  at (intersection-1);
    
    \path[name path=circle] (S) circle(2mm);
    
    \path [name intersections={of = circle and GStoFE_wb_top_left}];
    \coordinate (I1)  at (intersection-1);
    \coordinate (I2)  at (intersection-2);
    
    \node[empty] (I1_node) [below = 0.5 mm of I1] {};
    
    \draw [-] (GS.200) -| node[pos=0.5](GStoFE_wb_top_left) {} (I1_node);
    \draw [-] (I2) -| node[pos=0.5] {} (FE_wb_top_left);
    
    \draw[-] (I1.north) arc (90:278:2mm);
    
    \draw [->] (FE_wb_top_left.north) -| node[pos = 0.5] {$\alpha$} (FE_wb.north);
   
\end{tikzpicture}
    \caption{Schematic of the interaction force controller used in the simplified condition.}
    \label{fig:simp_control_diagram}
\end{figure*}

For the simplified controller, the legs are modeled as two-link chains independent of each other. This model is used to estimate interaction torques as in Equation~\ref{eq:simplified_interaction_estimation}. Since an optimization scheme considering physical limits is not implemented in the related works in Table~\ref{tab:related_work}, we do not use one for the simplified controller. Instead, Equation~\ref{eq:simpleEoM} is used as the forward model to track the desired acceleration commands sent by the virtual mass controller. The same controller gains for thigh and shank presented in Table~\ref{tab:vm_values} are used. The simplified interaction torque estimation is used in the control loop; however, the presented interaction torque results are calculated using the whole-body model to compare the results between conditions. The overall control structure is presented in Figure~\ref{fig:simp_control_diagram}. The control structure runs independently for each leg.

\subsection{Calculation of the Dynamic Parameters}
\label{app:dyn_params}
The mass matrix $\mt{M}$, Coriolis vector $\mt{b}$, and gravitational vector $\mt{g}$ are calculated during single stance in both WECC and the simplified model as
\begin{align}
    {\mt{M}_\text{i}}^{} & = \sum_{n \in \mt{N}_i}^{} {{}^{}_{}\mt{J}^{n}_{\text{S, i}}}^\top m_n {{}^{}_{}\mt{J}^{n}_{\text{S, i}}} + 
    {{}^{}_{}\mt{J}^{n}_{\text{R, i}}}^\top \boldsymbol{\Theta}_n {{}^{}_{}\mt{J}^{n}_{\text{R, i}}}, \\
    {\mt{b}_\text{i}}^{} & = \sum_{n \in \mt{N}_i}^{} {{}^{}_{}\mt{J}^{n}_{\text{S, i}}}^\top m_n {{}^{}_{}\dot{\mt{J}}^{n}_{\text{S, i}}} \dot{\mt{q}} + {{}^{}_{}\mt{J}^{n}_{\text{R, i}}}^\top \boldsymbol{\Theta}_n {{}^{}_{}\dot{\mt{J}}^{n}_{\text{R, i}}} \dot{\mt{q}}, \\
     {\mt{g}_\text{i}}^{} & = -\sum_{n \in \mt{N}_i}^{} {{}^{}_{}\mt{J}^{n}_{y,\text{ i}}}^\top m_n g , 
     \label{eq:g_i}
\end{align}
where $i \in \{\text{ls, rs, st, sw}\}$. The subscripts $i = \text{ls}$ and $i = \text{rs}$ correspond to left stance and right stance for the whole-body model, and $i = \text{st}$, $i = \text{sw}$ correspond to the state of a single leg being stance or swing for the simplified model as explained in Section~\ref{ssec:dynamic_model}. The index $n$ corresponds to the links of the exoskeleton in the following order: backpack ($n=1$), stance thigh ($n=2$), stance shank and foot ($n=3$), swing thigh ($n=4$), and swing shank and foot ($n=5$). For the whole-body model, all links are considered; therefore, $\mt{N}_{\text{ls}} = \mt{N}_{\text{rs}} = \{1,2,3,4,5\}$. On the other hand, the simplified model considers the legs separately; therefore, $\mt{N}_{\text{st}} = \{2,3\}$ and $\mt{N}_{\text{sw}} = \{4,5\}$. The mass and mass moment of inertia at the center of mass of each link are represented by $m_n$ and $\boldsymbol{\Theta}_n$, respectively. The translational and rotational Jacobian of the center of mass of each link are shown by ${{}^{}_{}\mt{J}^{n}_{\text{S, i}}}$ and ${{}^{}_{}\mt{J}^{n}_{\text{R, i}}}$, respectively. It is important to note that these Jacobians are calculated depending on the model and gait state (i.e., $i$). For the whole-body model, ${{}^{}_{}\mt{J}^{n}_{\text{S, ls-rs}}} \in \mathbb{R}^{2 \times 5}$ and ${{}^{}_{}\mt{J}^{n}_{\text{R, ls-rs}}} \in \mathbb{R}^{1 \times 5}$. For the simplified model, ${{}^{}_{}\mt{J}^{n}_{\text{S, st-sw}}} \in \mathbb{R}^{2 \times 2}$ and ${{}^{}_{}\mt{J}^{n}_{\text{R, st-sw}}} \in \mathbb{R}^{1 \times 2}$. The variable ${{}^{}_{}\mt{J}^{n}_{y, \text{i}}}$ is the vertical component of the translational Jacobian.

\subsection{Difference Between Interaction Torque Estimation Using Simplified and WECC Models}
\label{app:differences}
Due to the weight of the ExoMotus-X2 exoskeleton, gravitational torques $\mt{g}$ have the largest contribution to the interaction torque. Therefore, the difference of $\mt{g}$ between the whole-body and simplified model is investigated in this subsection.


The respective subsets of the whole-body Jacobians of the stance and swing center of masses are equal to the simplified stance and swing Jacobians,
\begin{equation}
\begin{split}
{{}^{}_{}\mt{J}^{n}_{y, \text{i}}}[2:3] &= {{}^{}_{}\mt{J}^{n}_{y, \text{st}}},  \;\:
    \;\: \forall n \in \{2,3\},
    \;\: i\in \{\text{ls}, \text{rs}\}, \\
{{}^{}_{}\mt{J}^{n}_{y, \text{i}}}[4:5] &= {{}^{}_{}\mt{J}^{n}_{y, \text{sw}}},  \;\:
    \;\: \forall n \in \{4,5\},
    \;\: i\in \{\text{ls}, \text{rs}\},
\end{split}
\label{eq:J_yi}
\end{equation}
where the notation [a:b] denotes the subset of elements from index a (inclusive) to index b (inclusive).

Using \eqref{eq:g_i} and \eqref{eq:J_yi}, the difference between the gravitational torques on the swing leg in the whole-body model and the simplified swing model is
\begin{equation}
{\mt{g}_\text{i}}^{}[4:5] - {\mt{g}_\text{sw}}^{} = \sum_{n \in {1,2,3}}^{} - {{}^{}_{}\mt{J}^{n}_{y, \text{i}}}[4:5] m_n g, \;\: i\in \{\text{ls}, \text{rs}\}.
\end{equation}

The state of the swing leg or backpack does not affect the center of mass positions on the stance leg with respect to the stance ankle. Therefore,
\begin{equation}
{{}^{}_{}\mt{J}^{n}_{y, \text{i}}}[4:5] = 0_{1 \times 2},  \;\:
    \;\: \forall n \in \{1,2,3\}, i\in \{\text{ls}, \text{rs}\}.
\end{equation}
This leads to
\begin{equation}
\begin{split}
 &\sum_{n \in {1,2,3}}^{} - {{}^{}_{}\mt{J}^{n}_{y, \text{i}}}[4:5] m_n g = 0, \\
 &\implies {\mt{g}_\text{i}}^{}[4:5] = {\mt{g}_\text{sw}}^{},  \;\: i\in \{\text{ls}, \text{rs}\},
\end{split}
\end{equation}
which shows the gravitational torques of the swing leg for the whole-body model are equivalent to the gravitational torques of the simplified swing model.

On the other hand, using \eqref{eq:g_i} and \eqref{eq:J_yi},  the difference between the gravitational torques on the stance leg in the whole-body model and the simplified stance model is
\begin{equation}
{\mt{g}_\text{i}}[2:3] - {\mt{g}_\text{st}} = \sum_{n \in {1,4,5}}^{} -{{}^{}_{}\mt{J}^{n}_{y, \text{i}}[2:3]}^\top m_n g  \;\: i\in \{\text{ls}, \text{rs}\},
\end{equation}
which shows the simplified stance gravitational torque has an error proportional to the weight of the backpack and swing legs. For heavy exoskeletons such as the ExoMotus-X2, this error results in significant inaccuracy in interaction torque estimation. 

\subsection{Double Stance Equation of Motion}
\label{app:ds}
Equation~\eqref{eq:DS2} can be written as an extension of both left stance dyanmics ($i=\text{ls}$) or right stance dynamics ($i=\text{rs}$). Therefore,
\begin{equation}
    \mt{H}^\top \mt{X}_{\text{ls}} = \mt{H}^\top \mt{X}_{\text{rs}},
    \label{eq:ls_rs}
\end{equation}
where $\mt{X}\in \{\mt{M}\ddot{\mt{q}}, \mt{b}, \mt{g}\}$. Multiplying Equations~\eqref{eq:DSEoM1}-\eqref{eq:DSEoM3} by $\mt{H}^\top$ on the left,

\begin{equation}
\begin{split}
\mt{H}^\top\mt{X}_{\text{ds}} &= \mt{H}^\top(\alpha \mt{X}_{\text{ls}} + (1-\alpha) \mt{X}_{\text{rs}}) \\
&= \mt{H}^\top\alpha\mt{X}_{\text{ls}} + \mt{H}^\top\mt{X}_{\text{rs}} - \mt{H}^\top\alpha\mt{X}_{\text{rs}} \\
&= \alpha(\mt{H}^\top\mt{X}_{\text{ls}} - \mt{H}^\top\mt{X}_{\text{rs}}) + \mt{H}^\top\mt{X}_{\text{rs}} \\
&= \mt{H}^\top\mt{X}_{\text{rs}} = \mt{H}^\top\mt{X}_{\text{ls}}.
    \label{eq:prf1}
\end{split}
\end{equation}

Therefore, the approximated dynamics of Equations~\eqref{eq:DSEoM1}-\eqref{eq:DSEoM3} satisfy the equation of motion~\eqref{eq:DS2}.

The interpolation factor, $\alpha$, is chosen based on the torques needed to carry the weight of the exoskeleton during the double stance. Let $\hat{\mt{g}}_\text{fly} \triangleq \mt{g}_\text{fly}[3:7]$ and $\hat{\mt{J}}_{\text{i}, y}  \triangleq \mt{J}_{\text{i}, y}[3:7]$. Indexes from 3 to 7 correspond to the generalized coordinates during single and double stance (i.e., $\theta_{0-4}$). The variable $\mt{J}_{\text{i}, y}$ is the vertical ($y$) component of the left ($i = \text{l}$) or right ($i = \text{r}$) ankle Jacobian in flight state.

The gravity vector during left and right stance can be expressed by the sum of the gravity vector during the flight state and the reflected joint torques due to the force on the stance foot,
\begin{equation}
\begin{split}
\mt{g}_{\text{ls}} &= \hat{\mt{g}}_\text{fly} - \hat{\mt{J}}_{\text{l}, y}^\top m_{\text{exo}}g \\
\mt{g}_{\text{rs}} &= \hat{\mt{g}}_\text{fly} - \hat{\mt{J}}_{\text{r}, y}^\top m_{\text{exo}}g,
\label{eq:g_ls_rs}
\end{split}
\end{equation}
where $m_{\text{exo}}g$ is the total weight of the exoskeleton.

Similarly, the gravity vector during double stance is equal to the sum of the flight gravity vector and the reflected joint torques due to the vertical ground reaction forces on both legs,
\begin{equation}
\begin{split}
\mt{g}_{\text{ds}} &= \hat{\mt{g}}_\text{fly} - \hat{\mt{J}}_{\text{l}, y}^\top F_{\text{l}, y} - \hat{\mt{J}}_{\text{r}, y}^\top F_{\text{r}, y} \\
&= \alpha\mt{g}_\text{ls} + (1-\alpha)\mt{g}_\text{rs}.
\label{eq:g_ds}
\end{split}
\end{equation}
Substituting \eqref{eq:g_ls_rs} into \eqref{eq:g_ds},
\begin{equation}
\begin{split}
&\hat{\mt{g}}_\text{fly} - \hat{\mt{J}}_{\text{l}, y}^\top F_{\text{l}, y} - \hat{\mt{J}}_{\text{r}, y}^\top F_{\text{r}, y} = \\
&\alpha \hat{\mt{g}}_\text{fly} - \alpha \hat{\mt{J}}_{\text{l}, y}^\top m_{\text{exo}}g + \hat{\mt{g}}_\text{fly} - \hat{\mt{J}}_{\text{r}, y}^\top m_{\text{exo}}g - \alpha \hat{\mt{g}}_\text{fly} + \alpha \hat{\mt{J}}_{\text{r}, y}^\top m_{\text{exo}}g.
\label{eq:g_ds_2}
\end{split}
\end{equation}
Rearranging and simplifying \eqref{eq:g_ds_2} leads to
\begin{equation}
\begin{split}
\alpha m_{\text{exo}}g (\hat{\mt{J}}_{\text{r}, y}^\top - \hat{\mt{J}}_{\text{l}, y}^\top) = \hat{\mt{J}}_{\text{r}, y}^\top(m_{\text{exo}}g - F_{\text{r}, y}) - \hat{\mt{J}}_{\text{l}, y}^\top F_{\text{l}, y}.
\label{eq:g_ds_3}
\end{split}
\end{equation}
Assuming vertical acceleration is not significant, i,e., $m_{\text{exo}}g = F_{\text{l}, y} + F_{\text{r}, y}$ results in
\begin{equation}
\begin{split}
\alpha(F_{\text{l}, y} + F_{\text{r}, y})(\hat{\mt{J}}_{\text{r}, y}^\top - \hat{\mt{J}}_{\text{l}, y}^\top) &= F_{\text{l}, y}(\hat{\mt{J}}_{\text{r}, y}^\top - \hat{\mt{J}}_{\text{l}, y}^\top)\\
\implies \alpha &= \frac{F_{\text{l}, y}}{F_{\text{l}, y} + F_{\text{r}, y}}.
\label{eq:g_ds_4}
\end{split}
\end{equation}

\subsection{Supplementary Kinematic Data}

\begin{figure}
\centering
\hspace*{-0.3cm}
\includegraphics[height=0.25\textwidth]{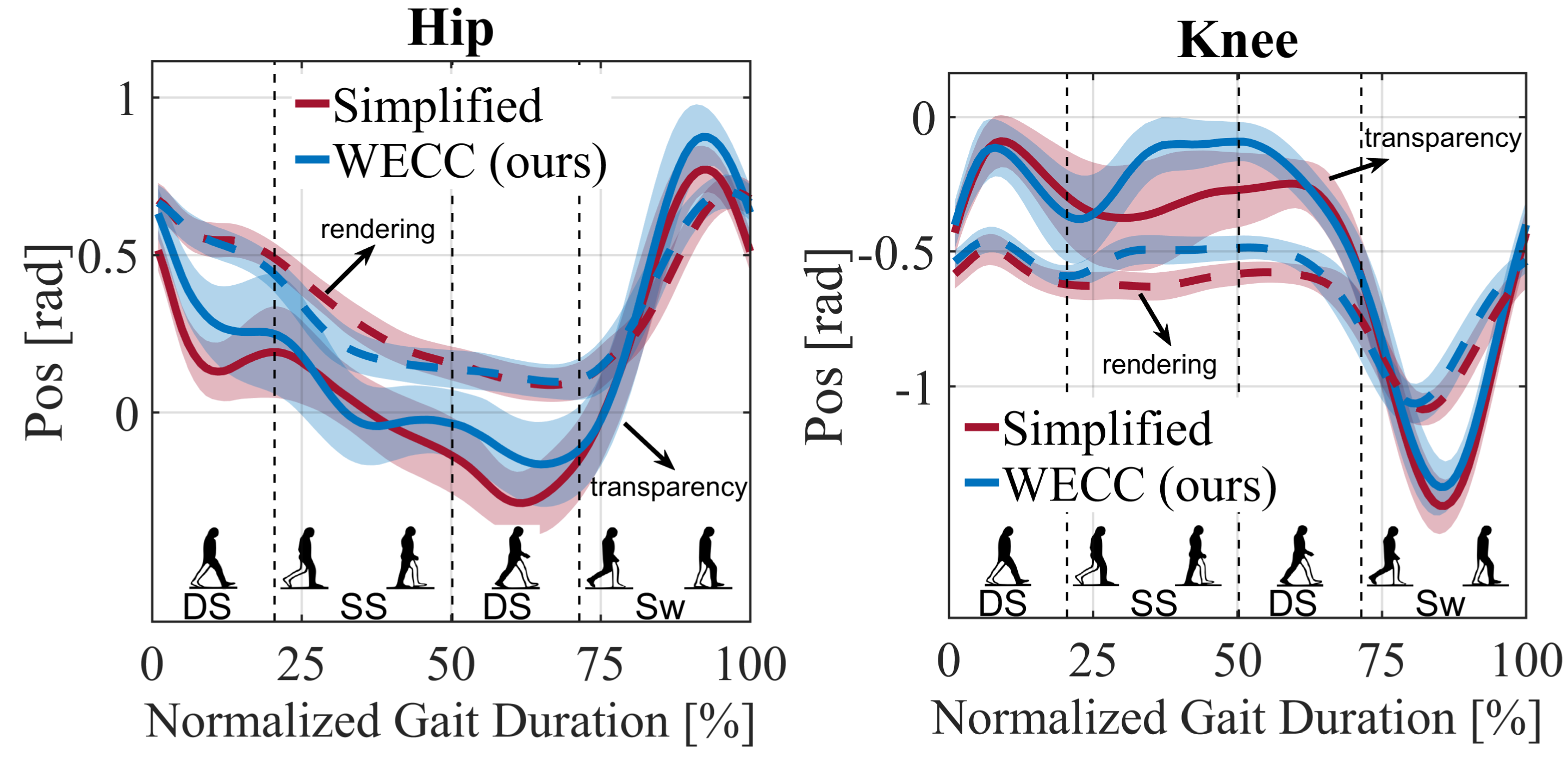}
\caption{Hip and knee joint positions for simplified and WECC controller during transparency (solid line) and haptic rendering experiments (dashed line) for a representative subject.}
\label{fig:joint_pos}
\end{figure}

Figure~\ref{fig:joint_pos} shows the joint positions observed during the transparency and haptic rendering conditions. In both conditions, similar joint trajectories were observed. However, during the haptic rendering trials, a significantly smaller range of motion was observed. This can be explained by the rendered torques due to the virtual spring and damper elements.

\color{black}

\addtolength{\textheight}{-0cm}   

\section*{Acknowledgment}
This work was supported by the National Science Foundation~/~National Robotics Initiative (Grant No: 2024488), Northwestern University and the Turkish Fulbright Commission. We would like to thank Tim Haswell for his technical support on the hardware improvements of the ExoMotus-X2 exoskeleton and Lorenzo Vianello for his help with the experimentation. \textcolor{black}{Presented data will be available upon request.}
\bibliographystyle{bibliography/myIEEEtran} 
\bibliography{bibliography/references}

\begin{IEEEbiography}[{\includegraphics[width=1in,height=1.25in,clip,keepaspectratio]{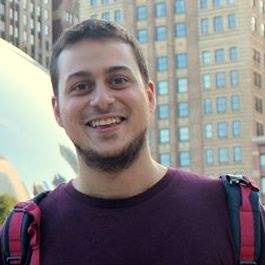}}]
{Emek Barış Küçüktabak} (S’21) received his B.S. degree in Mechanical Engineering from Middle East Technical University, Ankara, Turkey in 2017 and M.S. degree in Mechanical Engineering from ETH Zurich, Switzerland in 2019. Currently, he is a Ph.D. student at the Center for Robotics and Biosystems at Northwestern University and the Legs+Walking Lab at the Shirley Ryan AbilityLab, Chicago, IL. His research interests include physical human-robot interaction, teleoperation, force control, and assistive robotics.
\end{IEEEbiography}

\begin{IEEEbiography}[{\includegraphics[width=1in,height=1.25in,clip,keepaspectratio]{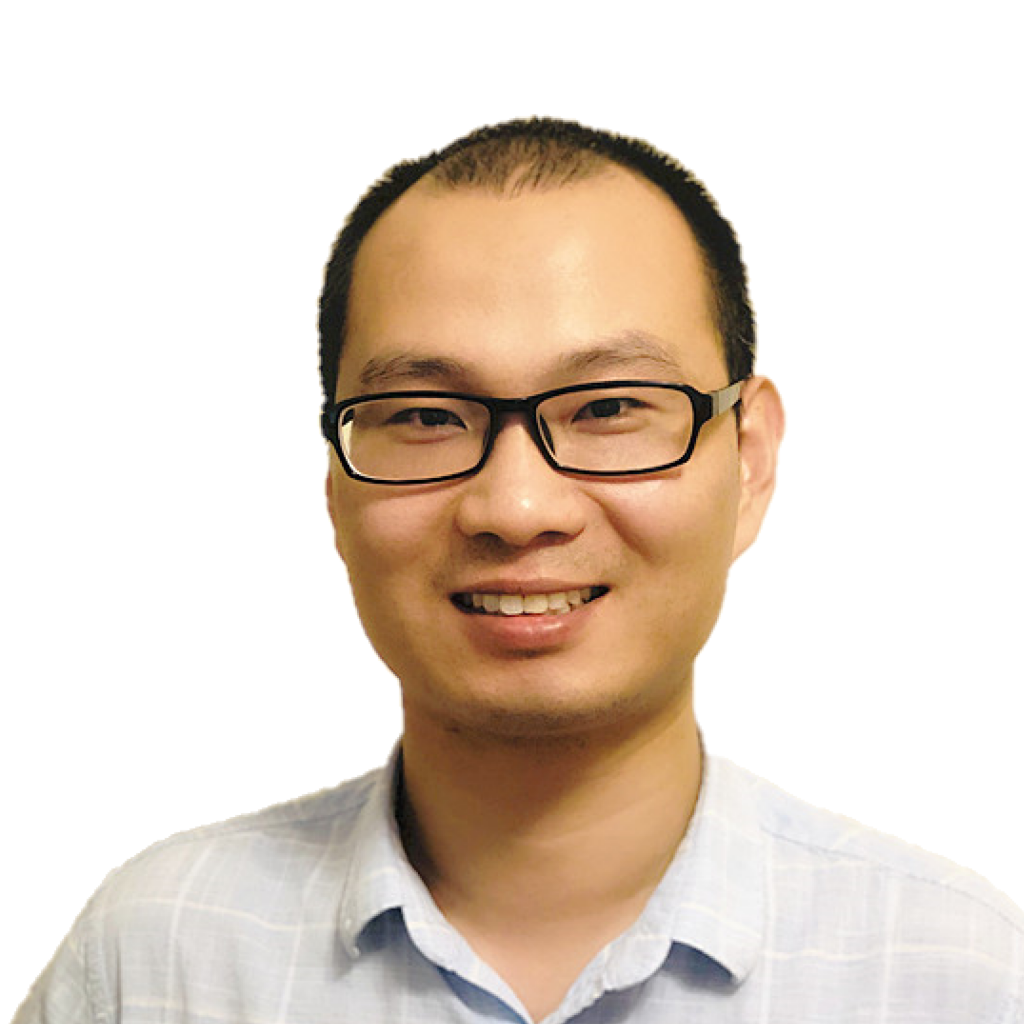}}]
{Yue Wen} received the B.S. degree in Automation from Wuhan University of Technology, China, in 2011, the M.S. degree in Control Theory and Engineering from Huazhong University of Science and Technology, China, in 2014, and the Ph.D. degree in Biomedical Engineering from North Carolina State University and the University of North Carolina at Chapel Hill, Raleigh, NC, in 2019. He is currently an Assistant Professor in the Department of Mechanical and Aerospace Engineering at the University of Central Florida, Orlando, FL. His research interests include the control and personalization of wearable robots, deep learning for neural machine interfaces, human-robot interaction, and gait rehabilitation.
\end{IEEEbiography}

\begin{IEEEbiography}[{\includegraphics[width=1in,height=1.25in,clip,keepaspectratio]{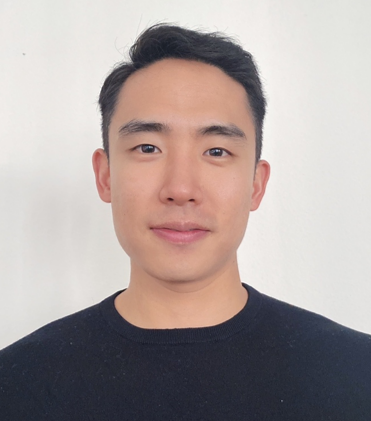}}]
{Sangjoon J. Kim} earned his B.S. from the Department of Electrical and Computer Engineering at the University of Wisconsin–Madison in 2012. Subsequently, he attained both his M.S. and Ph.D. degrees in Mechanical Engineering from the Korea Advanced Institute of Science and Technology (KAIST) in 2014 and 2019, respectively. Currently, he serves as an Assistant Project Scientist at the University of California-Irvine, concurrently holding a position as a Research Engineer at Flint Rehabilitation. His primary focus lies in biomedical engineering and robotics.  
\end{IEEEbiography}

\begin{IEEEbiography}[{\includegraphics[width=1in,height=1.25in,clip,keepaspectratio]{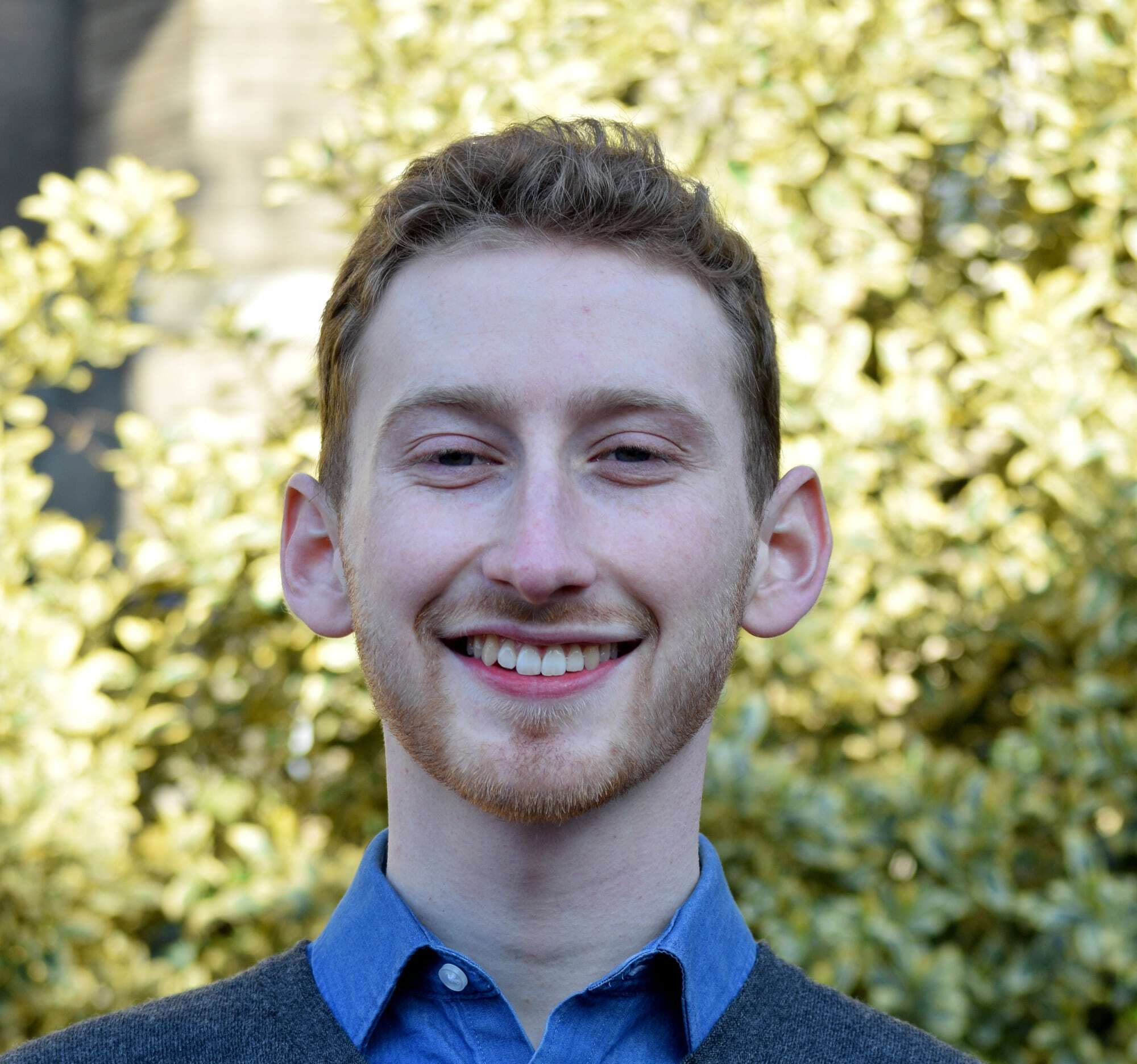}}]
{Matthew R. Short} received his B.S. degree in bioengineering from Temple University, Philadelphia, PA, USA in 2018. He is currently pursuing a Ph.D. degree at Northwestern University, Chicago, IL, USA, working in the Neurorehabilitation and Neural Engineering Lab at Shirley Ryan AbilityLab.
His research interests include human-human interaction and ankle motor control in the context of post-stroke rehabilitation.
  
\end{IEEEbiography}

\begin{IEEEbiography}[{\includegraphics[width=1in,height=1.25in,clip,keepaspectratio]{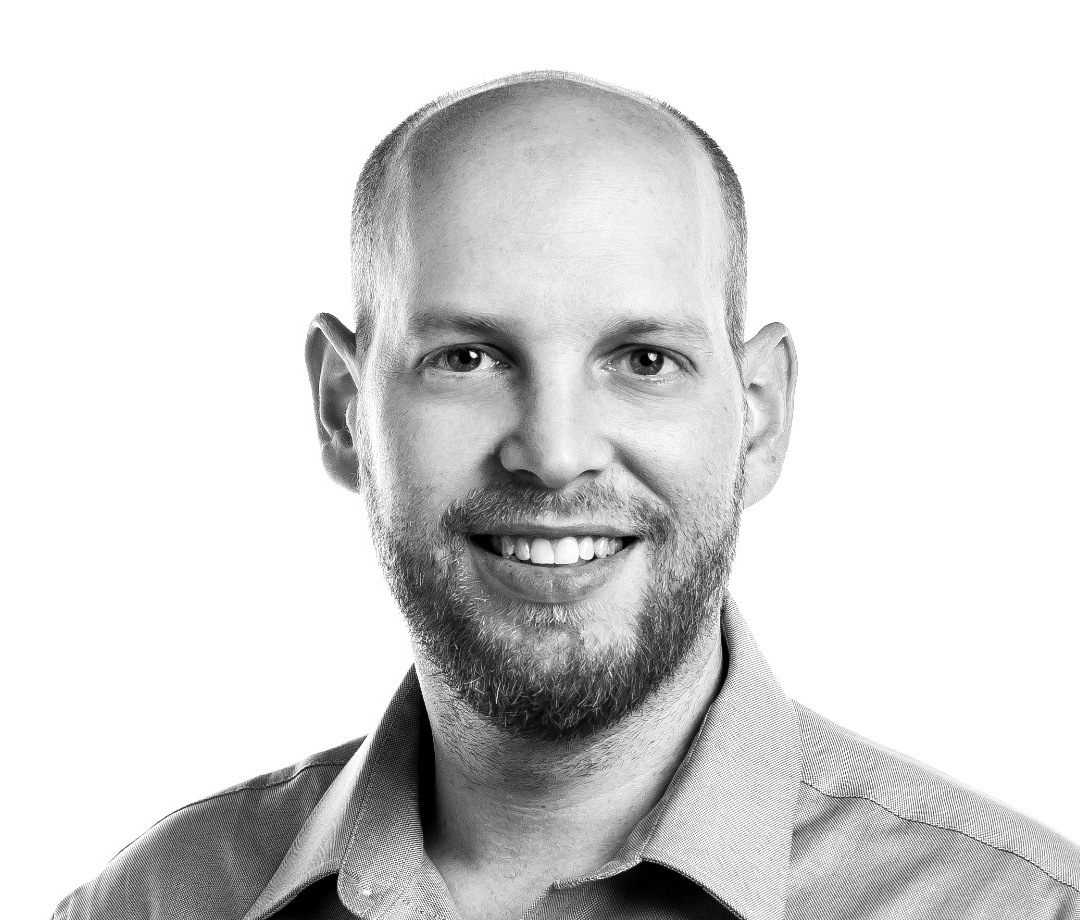}}]
{Daniel Ludvig} (S’06-M’11) received the B.Sc. degree in physiology and physics in 2003, and the M.Eng. and Ph.D. degrees in biomedical engineering in 2006 and 2010 respectively, all from McGill University, Montreal, QC, Canada. He is currently a Research Assistant Professor in the Department of Biomedical Engineering at Northwestern University and a Research Scientist at the Shirley Ryan AbilityLab in Chicago, IL. Dr. Ludvig specializes in using quantitative and experimentally driven techniques to model human biomechanics and motor control. His current specific research interests lie in quantifying the role lower-limb neuromechanics play in healthy locomotion, and how alterations in these neuromechanics lead to injury and disease.  
\end{IEEEbiography}

\begin{IEEEbiography}[{\includegraphics[width=1in,height=1.25in,clip,keepaspectratio]{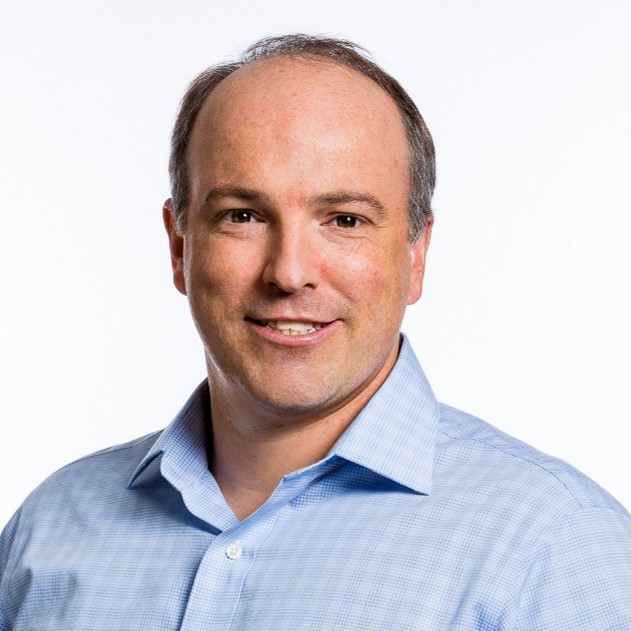}}]
{Levi Hargrove} received his MScE and PhD in Electrical Engineering from the University of New Brunswick (2005, 2008). He is currently the Director and Scientific Chair of Center for Bionic Medicine at the Shirley Ryan AbilityLab and an Associate Professor in the Departments of Physical Medicine \& Rehabilitation and the McCormick School of Engineering at Northwestern University. His research interests include signal processing, pattern recognition, and control of bionic limbs.   
\end{IEEEbiography}

\begin{IEEEbiography}[{\includegraphics[width=1in,height=1.25in,clip,keepaspectratio]{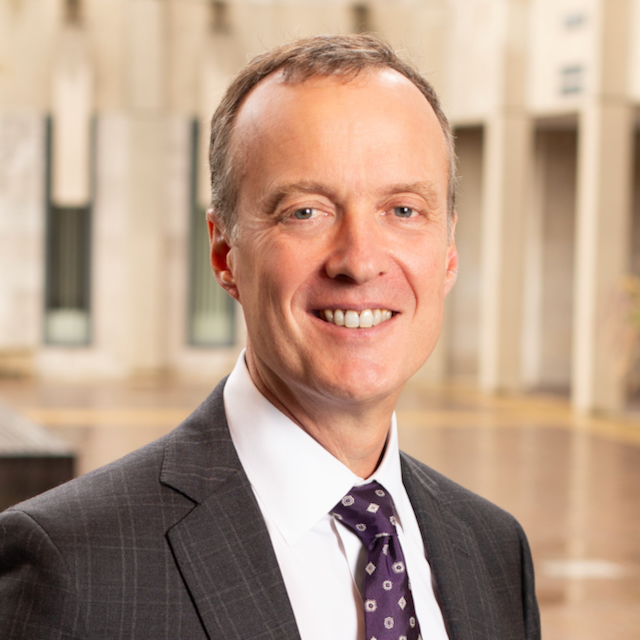}}]
{Eric J. Perreault} is Vice President for Research and Professor of Biomedical Engineering and Physical Medicine and Rehabilitation at Northwestern University. His research addresses the neural and biomechanical factors involved in the control of multi-joint movement and posture and how they are modified following neuromotor pathologies such as stroke and spinal cord injury. Eric is a fellow of the American Institute for Medical and Biological Engineering, and was recently chair of the National Advisory Board on Medical Rehabilitation Research, and president of the International Society for Electrophysiology and Kinesiology.  
\end{IEEEbiography}

\begin{IEEEbiography}[{\includegraphics[width=1in,height=1.25in,clip,keepaspectratio]{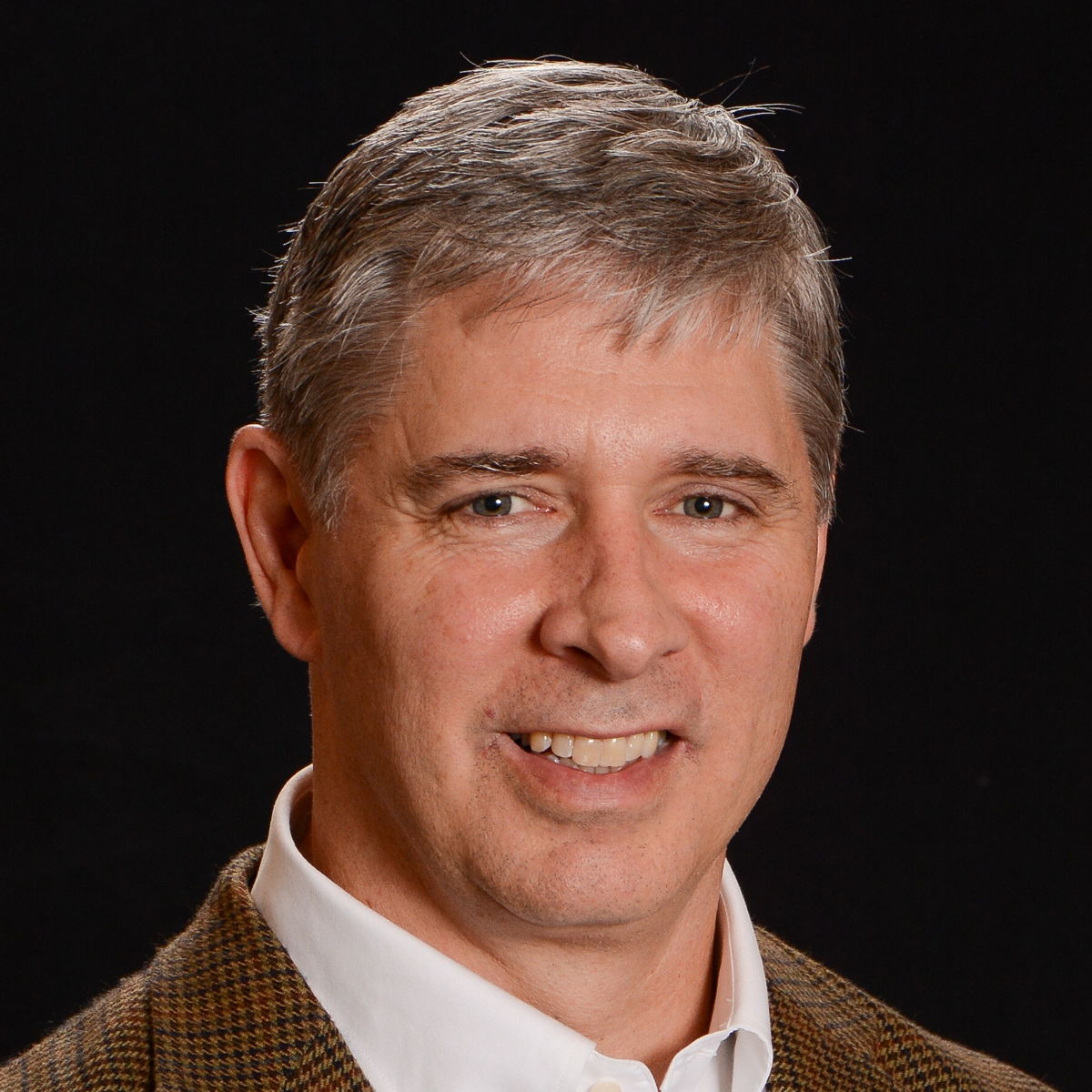}}]
{Kevin M. Lynch} (S’90–M’96–SM’05–F’10) received the B.S.E. degree in electrical engineering
from Princeton University, Princeton, NJ, USA, and the Ph.D. degree in robotics from
Carnegie Mellon University, Pittsburgh, PA, USA.

He is a professor of mechanical engineering at Northwestern University, where he directs the Center for Robotics and Biosystems and is a member of the Northwestern Institute on Complex Systems. He is a coauthor of the textbooks \emph{Principles of Robot Motion} (Cambridge, MA, USA: MIT Press, 2005) and \emph{Modern Robotics: Mechanics, Planning, and Control} (Cambridge, U.K.: Cambridge Univ. Press, 2017) and the associated online courses and videos. His research interests include robot manipulation and locomotion, self-organizing multiagent systems, and physically collaborative human–robot systems.  
\end{IEEEbiography}

\begin{IEEEbiography}[{\includegraphics[width=1in,height=1.25in,clip,keepaspectratio]{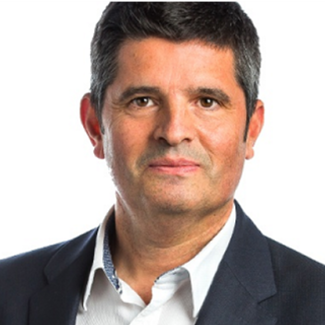}}]
{José L. Pons} graduated in Mechanical Engineering (Universidad de Navarra, Spain, 1992), and obtained a MSc in Information Technologies (Universidad Politécnica de Madrid, Spain, 1994), and a PhD in Physics (Universidad Complutense de Madrid, 1997), for his research work on dynamic optimization of robot mechanisms.

Dr. Pons currently serves as the Scientific Chair of the Legs \& Walking AbilityLab at Shirley Ryan AbilityLab. In addition to his research leadership role at SRAlab, he holds appointments as Professor at the Department of Physical Medicine \& Rehabilitation, Feinberg School of Medicine and at the Departments of Biomedical Engineering and Mechanical Engineering, McCormick School of Engineering, Northwestern University.
\end{IEEEbiography}










\end{document}